\ifdefined\XeTeXversion\else\pdfoutput=1\fi 
\documentclass[11pt,letterpaper]{article}

\usepackage[margin=1in]{geometry}
\usepackage{amsmath,amssymb}
\usepackage{booktabs,tabularx,array}
\usepackage{graphicx}
\usepackage{float}
\usepackage{microtype}
\usepackage[font=small,labelfont=bf]{caption}
\usepackage[round,authoryear]{natbib}
\usepackage{xcolor}
\usepackage{xurl}
\usepackage[hidelinks]{hyperref}

\graphicspath{{figures/}}
\newcommand{\Iidx}{I}

\hypersetup{
  pdftitle={A Glyph Is Not a Letter, a Token Is Not a Word, a Space Is Not a Space: What the Units of Voynichese Are Not},
  pdfauthor={Liudmila Rozanova and Alexander Temerev},
  pdfsubject={Voynich manuscript; analytical units; token order; boundary statistics},
  pdfkeywords={Voynich manuscript, analytical units, segmentation, token order, conditional entropy, byte-pair encoding}
}

\title{A Glyph Is Not a Letter, a Token Is Not a Word,\\
a Space Is Not a Space:\\
What the Units of Voynichese Are Not}
\author{%
{\large\bfseries Liudmila Rozanova\textsuperscript{1}, Alexander Temerev\textsuperscript{2}}\\[6pt]
\normalsize
\textsuperscript{1}International Institute for Applied Systems Analysis (IIASA), Laxenburg, Austria\\
\textsuperscript{2}University of Geneva, Geneva, Switzerland}
\date{August 2026}

\begin{document}
\maketitle

\begin{abstract}
The Voynich manuscript (Beinecke MS 408) is usually analysed on three
unstated assumptions: that its glyphs are letters, that the strings
between blanks are words, and that every blank is a word space. We test all
three against the Zandbergen--Landini transliteration with matched prose, cipher, and
pseudo-text controls and quire-level resampling. None holds, and the
failures share a shape: the order in Voynichese sits at the edges of tokens
and at graded boundaries between them, not in the succession of tokens
themselves. Glyph regularity is too strong for one-to-one substitution of
any tested plaintext (conditional entropy 2.7 bits against
about 3.5 for Latin, Italian, and English) and resolves instead onto a
quire-stable scale of recurrent multi-symbol units. Tokens form a plausible
vocabulary, yet the identity of one token predicts the next by under 1\% of
token entropy, below every matched control (2--10\%), while the glyphs at
token edges share 0.2 bits of mutual information, more than in any prose
control. Blanks fall into two regimes: the separators transcribers marked
uncertain behave like word-internal junctures, are physically narrower on the
page (AUC 0.905 from independent image coordinates, with the same sign in a
small blind ink audit), and are crossed by learned units even when every space is
erased before learning. This profile is also what discriminates. A published
Voynich-imitating cipher and a self-citation text generator both reproduce
the low entropy, the unit scale, the weak token order, and the null result of
a calibrated substitution attack; neither reproduces the edge-glyph coupling
or the open, hapax-rich vocabulary (70\% singleton types against 41\% and
59--60\%). Any account of the manuscript must therefore earn, rather than
assume, the step from glyphs, tokens, and separators to letters, words, and
word spaces, and these are the measurements on which to do so.
\end{abstract}


\section{Introduction}

Every decipherment begins by deciding what the signs and divisions are. In a
known writing system, terms such as \emph{letter}, \emph{word}, and \emph{word
space} carry linguistic interpretations. In the Voynich manuscript (Beinecke
MS 408), the direct observations are more modest: recurring marks, strings of
marks separated by visible gaps, and several transcription conventions. A
transcription makes those observations countable; it does not by itself turn
them into linguistic units \citep{beinecke,bowern2021}.

For operational purposes, quantitative work commonly makes three
identifications: an EVA glyph is counted as a plaintext letter, a string
between transcription separators is called a word, and every separator is
treated as the same word boundary \citep{landini2001,reddy2011,smith2019,
lindemann2021}. These choices are convenient, but each embeds part of the
decipherment in the input. The underlying unit problem is old: early
cryptanalytic discussions explicitly asked whether the visible groups might
represent letters, syllables, words, or mixed-length strings
\citep{currier1976}.
If a glyph belongs to a larger recurrent group, letter-level statistics are
measured at the wrong scale. If a token is not a lexical word, vocabulary and
word-order statistics are misnamed. If blanks mark more than one type of
juncture, pooling them destroys the very structure under investigation.

The central claim of this paper is therefore methodological and testable. A
transcription glyph is not yet a demonstrated letter; a blank-delimited token
is not yet a demonstrated word; and a separator is not automatically a word
boundary. We ask what the data permit each observed unit to mean. The three
tests and their warranted interpretations are summarised in
Table~\ref{tab:unit-assumptions}.

\begin{table}[H]
\centering
\small
\caption{Three common unit identifications tested in this paper. Rejection as
a default does not imply that no individual glyph, token, or separator can
serve the proposed function.}
\label{tab:unit-assumptions}
\begin{tabularx}{\textwidth}{>{\raggedright\arraybackslash}p{0.22\textwidth}
>{\raggedright\arraybackslash}p{0.35\textwidth}X}
\toprule
Assumed equivalence & Main tests & Warranted interpretation \\
\midrule
EVA glyph = plaintext letter & Conditional-entropy invariance; BPE scale;
calibrated cipher attack & Fixed one-to-one letter mapping is rejected for the
tested sources; recurrent multi-symbol units are the supported analytical scale. \\
ZL token = lexical word & Static vocabulary; shuffle-corrected token and
edge-glyph order; uncertain-space merging & The forms look lexical in
isolation but their identities do not carry the adjacent ordering expected of
tested prose or catalogue words; the order that exists sits at the token
edges. \\
Separator = uniform word boundary & Boundary association; image gaps;
space-erased BPE & Blanks are non-random but form at least two graded boundary
regimes; lexical status is not established. \\
\bottomrule
\end{tabularx}
\end{table}

For the glyph hypothesis, first-order conditional entropy supplies a formal
invariance test: a fixed one-to-one substitution changes symbol names but not
their probabilities or adjacent-pair structure. Byte-pair encoding (BPE) then
learns recurring multi-symbol units without naming them in advance
\citep{sennrich2016}. A calibrated class-merging attack asks whether the same
scale conceals ordinary prose under the tested verbose or homophonic ciphers.

For the word hypothesis, we separate lexical appearance from sequential
function. Vocabulary size and frequency concentration ask whether the isolated
forms look plausible. Adjacent-token mutual information, corrected by
within-line shuffling, asks whether those forms constrain their neighbours as
words do in matched prose and botanical records. Merging every uncertain
separator tests whether false splits explain the result.

For the space hypothesis, a transcription-only association score compares the
glyphs meeting at certain and uncertain separators. Label-independent image
coordinates then ask whether the classes correspond to physically different
gaps. Finally, we delete every space before learning units and test which
former positions are recovered as boundaries. This distinguishes the claim
that spaces are arbitrary from the stronger and different claim that all
spaces are lexical word breaks.

Much of the raw material is not new. Unusually low second-order character
entropy has been known since \citet{bennett1976}; dependencies between the last
glyph of one token and the first of the next since \citet{currier1976};
weak word-to-word predictability was reported by \citet{reddy2011} and
\citet{caruana2022}; non-uniform symbol roles, line-position effects,
directional asymmetries in the glyph sequence, and
word-structure grammars have been described repeatedly
\citep{dimperio1978,stolfi2000,landini2001,vogt2012,smith2019,lindemann2021,
matlach2022,zattera2022,feaster2021,steckley2024,bowern2021,parisel2025,
parisel2026a}; and both
encipherment and human-made or algorithmic pseudo-text can reproduce selected
word-level statistics \citep{rugg2004,schinner2007,timm2020,gaskell2022,
bowern2022,greshko2025}. Transcribers have also long noted that the glyph pairs
flanking uncertain spaces differ from those flanking ordinary spaces
\citep{feaster2020,robgea2022}. The
contribution here is to make these observations one falsifiable
unit-identification argument with matched controls and cluster-aware
uncertainty: we separate lexical appearance from sequential function,
distinguish separator classes before consulting the images and then validate
them against layout coordinates and ink, erase the spaces and relearn the
units, test whether the learned scale survives held-out quires, and calibrate a
substitution attack against synthetic ciphers, and then ask how much of the
resulting profile a published Voynich-imitating cipher and a self-citation
generator already reproduce. We do not claim that the manuscript lacks
language or meaning. We show that its observed symbols, tokens, and
separators cannot be promoted to letters, words, and uniform word spaces
without additional evidence, and we identify which of our measurements still
separate the manuscript from its best imitations.

\section{Data and methods}

\subsection{Transcription and analysis corpus}

The boundary analysis uses the Zandbergen--Landini (ZL) transliteration,
version 3b in IVTFF 2.0 format \citep{zl3b,landini2001}. ZL encodes certain separators as
\texttt{.} and uncertain separators as \texttt{,}. We retain primary paragraph
loci and require at least four clean tokens per line. Frequent EVA composites
(\texttt{cth}, \texttt{ckh}, \texttt{cph}, \texttt{cfh}, \texttt{ch},
\texttt{sh}, \texttt{iin}, \texttt{in}, and \texttt{ee}) are collapsed to
single symbols. We call this the \emph{composite-collapsed} representation.
The original EVA character string, in which those conventions remain
multi-character sequences, is the \emph{decomposed} representation.

The resulting boundary corpus contains 3,880 lines on 206 pages, nested in 50
bifolios and 16 quires. (The transliteration distinguishes 18 quires; the two
zodiac-foldout quires contain no qualifying paragraph lines.) It contributes
29,482 intra-line separators and 3,634 continuation line breaks. Paragraph
breaks are not included in the line-break category. Two quires, the biological
and the recipe sections, hold 47\% of the qualifying lines, so 16 top-level
clusters is an upper bound on the effective number of independent units.

Table~\ref{tab:representations} states which representation enters each test.
The two symbol counts differ because the entropy and BPE drivers use slightly
different conservative cleaning rules, not because the representation changes
inside either analysis. Type counts likewise depend on the cleaning rule:
the collapsed entropy stream contains 47 symbol types (25 with at least 100
occurrences), the boundary-index table 34, and the decomposed BPE stream 25.

\begin{table}[H]
\centering
\small
\caption{Observed representations used in the paper. ``Symbol'' is a
transcriptional unit, not an assumed plaintext letter or palaeographic atom.}
\label{tab:representations}
\begin{tabularx}{\textwidth}{>{\raggedright\arraybackslash}p{0.24\textwidth}
>{\raggedright\arraybackslash}p{0.29\textwidth}>{\raggedright\arraybackslash}X}
\toprule
Representation & Size in the main analysis & Uses \\
\midrule
Composite-collapsed EVA & 141,310 symbols; 47 observed types (25 frequent) & Boundary
association, coordinate alignment, and the primary entropy comparison \\
Decomposed EVA & 173,572 entropy symbols; 173,076 BPE symbols & Representation
sensitivity and data-driven unit learning \\
ZL surface tokens & 32,747 blank-delimited forms & Static vocabulary and
adjacent-token order \\
\bottomrule
\end{tabularx}
\end{table}

Throughout the paper, \emph{symbol} denotes an element of the stated EVA
representation, \emph{token} a string delimited by the ZL separators, and
\emph{separator} a transcribed juncture. We reserve \emph{letter}, \emph{word},
and \emph{word boundary} for interpretations that require additional evidence.
The distinction is not new to the field: the informal term \emph{vord}, used
on the Voynich Ninja forum for a blank-delimited Voynich string as opposed to
a plaintext word \citep{voynichninja2020}, marks the same reservation at the
token level. What we add is to state the reservation at all three levels, to
test each with matched controls, and to say what the observed unit does
warrant.

\subsection{Token inventory and adjacent order}

The token-order analysis uses the 3,950 paragraph lines for which the complete
separator sequence is recoverable, giving 32,747 observed tokens. We analyse
both the transcribed tokenisation and a conservative alternative that merges
across every ZL-uncertain separator. Static diagnostics are the number of
distinct forms, the share of types seen once, and the fraction of all tokens
accounted for by the 50 most frequent forms.

Adjacent-token mutual information asks how much knowing token \(T_t\) reduces
uncertainty about \(T_{t+1}\) \citep{cover2006}. Direct plug-in estimates are
upward-biased when the vocabulary is large \citep{paninski2003}. We therefore
retain the 2,000 most frequent forms,
map the remainder to one \texttt{<other>} class, and estimate
\begin{equation}
 I_{\mathrm{excess}}=I(T_t;T_{t+1})-
 \mathbb{E}_{\mathrm{within\mbox{-}line\ shuffle}}
 [I(T_t;T_{t+1})].
\end{equation}
The null independently permutes token order inside each line 100 times. It
preserves token frequencies, line membership, and line lengths while removing
the observed local sequence. We report \(I_{\mathrm{excess}}\) in bits and as
a share of capped token entropy, \(I_{\mathrm{excess}}/H(T)\). The latter says
what percentage of uncertainty in the next form is accounted for by adjacent
ordering beyond the finite-sample baseline. Vocabulary caps of 250, 500,
1,000, 2,000, and 4,000 test sensitivity to the chosen resolution.

The continuous-text controls comprise narratives in Latin, English, French,
German, and Italian; Latin medical prose (Celsus); Latin botanical prose
(Pliny); and an English herbal (Culpeper). Each is cut to exactly 32,747 tokens
and wrapped to the observed Voynich line-length sequence. A ninth control uses
entries from Linnaeus's \emph{Species Plantarum}; it is truncated to the same
token count but keeps its own record boundaries. The classical Latin files are
Tesserae/Perseus electronic texts \citep{coffee2013,perseus}; the remaining
controls are identified by work and ebook number in the Project Gutenberg
corpus citation \citep{gutenbergcontrols}. This separates a catalogue or
register from running prose and tests domain as well as language rather than
treating one genre as universal. Stability checks analyse Currier A and B
separately \citep{currier1976} and omit each quire in turn. The statistic
constrains ordering at the transcribed-token scale; low adjacent information
is not, by itself, a test of meaning.

\subsection{Normalised internality index}

For a glyph pair \((a,c)\), define a pointwise-mutual-information-style
association score from within-token bigrams \citep{church1990},
\begin{equation}
s(a,c)=\log_2\frac{p_{\mathrm{int}}(a,c)}
{p_{\mathrm{int}}(a)p_{\mathrm{int}}(c)}.
\end{equation}
The joint distribution uses add-0.5 smoothing on the fixed collapsed-glyph
vocabulary. The two marginals are the same position-independent within-token
unigram distribution; this avoids treating token-final and token-initial
position preferences as evidence of association.

For boundary class \(b\), let \(\overline{s}_b\) be the mean score of the glyph
pairs meeting at that class. The normalised index is
\begin{equation}
\Iidx(b)=\frac{\overline{s}_b-\overline{s}_{\mathrm{random}}}
{\overline{s}_{\mathrm{internal}}-\overline{s}_{\mathrm{random}}},
\end{equation}
where \(\overline{s}_{\mathrm{internal}}\) is the mean over observed
within-token bigrams and \(\overline{s}_{\mathrm{random}}\) is the expectation
under the independent product of within-token unigram frequencies. Thus
\(\Iidx=1\) is the within-token anchor and \(\Iidx=0\) is the
independent-pair anchor. The index is an association measure, not a probability
that a separator is internal to a lexical word.

The certain/uncertain and positional classifications overlap. First-of-line
and mid-line rows therefore do not partition the certain/uncertain rows.
Continuation line breaks are scored using the final glyph of one line and the
initial glyph of the next line on the same page.

\subsection{Cluster-aware uncertainty}

Lines from the same page, bifolio, and quire share material and production
context. The headline intervals therefore resample the 16 quires with
replacement and recompute the unigram table, bigram table, anchors, and
boundary means in every bootstrap draw \citep{efron1993}. We use 1,500 draws
and fixed seeds.
These are stability intervals over the observed top-level units, not estimates
from a population of independent manuscripts. Leave-one-quire-out and
quire-wise sign analyses provide additional small-cluster checks.

An edge-composition null preserves the observed left-edge and right-edge glyph
distributions of each boundary class while breaking their pairing. It asks how
much of a boundary score is attributable to the marginal glyph preferences at
the two edges.

\subsection{Currier strata and line-start decomposition}

Lines are assigned to Currier A or B by the ZL page variable
\citep{currier1976,parisel2026b}, and within each stratum every table, anchor,
and boundary mean is recomputed from that stratum alone. The A\,--\,B
difference of the uncertain index is bootstrapped by resampling each
stratum's quires independently (1,500 draws) and compared with a
pre-specified \(\pm0.05\) equivalence margin. For the line-start
decomposition we compute, within a set of lines, the Jensen--Shannon
divergence between the distribution of line-initial glyphs and that of
mid-line token-initial glyphs, separately for lines carrying the IVTFF
paragraph-start mark and for all others, with a relabelling null and a
quire bootstrap of the difference (Appendix~\ref{app:dialect}).

\subsection{Image-coordinate validation}

The site voynichese.com provides per-token bounding boxes laid over Beinecke
page images under a Takahashi--Stolfi EVA reading \citep{voynichese,beinecke}.
The public archive snapshot used here is commit
\texttt{66f8adaadc120f93e8a4c906685ba66a9d3ab847} and contains files for 223
folios. These coordinates do not encode the ZL certain/uncertain labels,
although they do embody a second tokenisation and are therefore
label-independent rather than segmentation-independent. The boxes are integer
coordinates snapped to a 3-pixel grid on a layout roughly 600 pixels wide, so
a typical token is 15--90 pixels wide and gaps are quantised in steps of about
one fifteenth of a token width. Their construction protocol is not published,
so they are treated as a coarse corroborating layout measurement, not as
ground truth.

For consecutive boxes on a line, the horizontal gap is
\begin{equation}
g_i=\frac{x_{i+1}-(x_i+w_i)}{\operatorname{median}_{j\in f}(w_j)},
\end{equation}
where \(f\) is the folio. Negative values denote overlapping boxes. The visual
and ZL token sequences are aligned on collapsed forms using exact matching
blocks. A certainty label is transferred only when both flanking tokens match
and the visual boxes are on the same line.

The primary comparison is the certain-minus-uncertain difference in mean box
gap with a folio-clustered bootstrap. A fixed-effect model
\begin{equation}
g_i=\beta\,\mathbb{1}(\text{uncertain}_i)+\alpha_{\text{glyph pair}(i)}+\epsilon_i
\end{equation}
controls for the identity of the two flanking glyphs. Its interval also
resamples folios. A two-way specification additionally absorbs folio fixed
effects. We quantify discrimination by the rank area under the ROC curve
(AUC), interpreted here as the probability that a randomly selected certain
gap exceeds a randomly selected uncertain gap \citep{fawcett2006}; its interval
resamples folios.
For an out-of-folio check, a single gap threshold maximising balanced accuracy
is fitted on all other folios and applied to the held-out folio. This is a
validation diagnostic, not a proposed retranscription algorithm. The pooled
comparison, fixed effects, and AUC are primary; the correlation between
continuous association score and box gap is secondary.

We add a second-reader check using Glen Claston's v101 transcription
\citep{v101}. Lines are compared only when folio, line number, and separator
count agree. We report raw agreement and Cohen's \(\kappa\) for certain versus
uncertain marks \citep{cohen1960}. For positions also aligned to the coordinate boxes, we then
group physical gaps by whether zero, one, or both transcriptions mark the
separator uncertain. This check is independent of the ZL labeller but still
shares the same underlying manuscript images and cannot establish lexical
status.

\subsection{Conditional-entropy comparison}

The entropy analysis uses all 34,175 clean tokens in primary paragraph loci.
For both the composite-collapsed and decomposed streams we report marginal
entropy \(H(X)\) and first-order conditional entropy
\(H(X_{t+1}\mid X_t)\). Spaces and line breaks are removed rather than treated
as symbols \citep{cover2006}. The composite-collapsed stream is primary because it treats common
multi-character EVA conventions as single glyph-like units; decomposed EVA is
a representation sensitivity. The controls are botanical
Latin (Pliny), other Latin prose (Celsus and Caesar), an English herbal
(Culpeper), and Italian prose. Each is truncated to the Voynich token count and
processed with the same plug-in estimator.

Two sensitivity analyses remove sample-size and boundary-concatenation
explanations. First, we draw 300 contiguous windows from each full control,
each exactly matching the 141,310-symbol Voynich stream. Second, we recompute
conditional entropy using only adjacent symbols inside transcribed tokens,
excluding every cross-token transition. We also report conditional
perplexity, \(2^{H(X_{t+1}\mid X_t)}\), as the effective number of next-symbol
choices.

The comparison tests a precise invariance. If each plaintext letter is mapped
bijectively to one symbol in either fixed representation, then relabeling
leaves both marginal and conditional entropy unchanged \citep{cover2006}. The controls do not
span every language, register, or writing system, so the result constrains only
source texts with comparable first-order structure.

\subsection{Learned-unit scale comparison}

We run BPE \citep{sennrich2016} on 173,076 decomposed EVA symbols from the primary paragraph loci. Merges
are learned within transcribed tokens and are not allowed to cross a separator.
At merge count \(k\), the resulting unit stream is \(U_k\). We measure adjacent
dependence as
\[
D_k=H(U_k)-H(U_{k,t+1}\mid U_{k,t}),
\]
the dependence gap between marginal and conditional uncertainty at that scale. We
evaluate \(k=0,4,8,16,32,64,128,256,512,\) and \(1{,}024\). Unit inventory,
occurrence count, mean glyph span, and \(D_k/H(U_k)\) are reported alongside
the gap.

Controls use the same Latin prose and approximately 173,000 output symbols:
plain Latin, two deterministic verbose substitutions (fixed two-glyph groups
and frequency-ranked one-to-three-glyph groups), and a homophonic verbose
substitution with three weighted two-glyph variants per plaintext letter.
These controls calibrate the scale curves; resemblance to one control is not
treated as model identification.

Because a BPE trough is not cipher-specific, we also use the bundle's
calibrated substitution attack. After BPE at \(k=64\), a Brown-style
distributional merge \citep{brown1992} reduces the learned units to 23 classes (sensitivities:
20 and 26). A near-bijective class-to-letter mapping is then optimised against
character-trigram language models and evaluated on held-out lines, following
the general language-model approach to substitution and homophonic-cipher
decipherment \citep{ravi2011}. The main
models are Latin, vowelless Latin, Italian, German, French, English, and
consonantal Hebrew, fitted separately to Currier A and B; Hebrew is included
because an earlier decipherment-style search ranked it first
\citep{hauer2016}, and the corpus was retrieved through the Sefaria API
\citep{sefaria}. An order-3 glyph
generator fitted to each Voynich stream supplies a matched negative control
with the same token and line shapes. End-to-end calibration uses five
synthetic ciphers of the same Latin: two, three, or five homophones per
letter, a variable one-to-three-symbol condition, and a three-homophone
cipher with rotating tables. The class count equals the plaintext alphabet
size (22 for the Latin synthetics). We report the real-minus-surrogate
language-match differential, calibrated over five cipher seeds per
configuration against both the right-language and a wrong-language model.
Null insertion, transposition, larger groups, and state-dependent encodings
are not included in this calibration.

Stability is tested five ways. We fit Currier A and B separately; refit after
omitting each of the 16 quires; and form 100 random partitions of the quires
into disjoint halves, producing 200 half-manuscript fits. At \(k=64\) we compare
the learned compound-unit sets across the two disjoint halves. To remove the
circularity of learning and scoring a segmentation on the same text, a
cross-fit learns the ordered BPE rules on 15 quires, applies them without
refitting to the omitted quire (including unseen word types), and scores only
that held-out quire. Fold statistics are averaged with held-out glyph count as
weight. Finally, we repeat the Voynich curve after collapsing nine standard
multi-character EVA composites before BPE, asking whether the scale pattern
merely reverses a known transcription convention. Stability refits use
\(k=0,16,32,64,128,\) and \(256\).

\subsection{Separator-erasure sensitivity}

To test whether imposed tokenisation creates the BPE trough, we retain the
3,950 paragraph lines with an exactly recoverable separator sequence, erase
all 28,797 intra-line separators, and fit BPE to each continuous line string.
This subset contains 166,280 EVA glyphs. We record whether each original
separator position lies inside a learned unit or at a learned-unit boundary.
The same calculation is applied to Latin and two verbose-cipher controls after
their genuine plaintext word spaces are hidden.

Two randomisation checks keep the line strings unchanged. The first places the
same number of separators uniformly at random within every line. The second
permutes each line's observed token lengths, preserving exactly its number and
multiset of lengths while changing their alignment with the glyph stream.
Each check uses 50 fixed-seed repetitions. Finally, for every adjacent symbol
position we predict space versus no space from the flanking decomposed-symbol pair.
Pair probabilities are trained on 15 quires and evaluated on the held-out
quire with a five-observation marginal backoff, cycling through all 16 quires.

\subsection{Edge-glyph order, two further controls, and published inventories}

Token-identity information is blind to structure carried by the glyphs at
token edges, where \citet{currier1976}, \citet{reddy2011}, and
\citet{parisel2026a} located the cross-space dependencies of Voynichese. On
the same corpora, line template, cap, and shuffle null as the token-order
analysis we therefore also estimate the shuffle-corrected information between
the last glyph of one token and the first of the next (and three related
pairings), each as a share of the marginal entropy of the predicted variable,
together with the fraction of adjacent tokens within one edit of each other,
the adjacent-similarity diagnostic of \citet{timm2020}. Two published
accounts that address the whole profile are then run through the same
drivers: the Naibbe cipher of \citet{greshko2025}, a hand-workable verbose
homophonic cipher that renders Latin as Voynich-like text, re-implemented
from its published tables and applied to the Latin control; and the
copy-and-modify self-citation generator of \citet{timm2020}, implemented from
their description and calibrated only to the Voynich mean token length and
type--token ratio. Both receive the full battery---glyph entropy, the BPE
curve, token-order and edge-glyph information, and the calibrated attack
differential against their own order-3 surrogates
(Appendix~\ref{app:controls}). Finally, the units learned at 32 and 64 merges
are compared with the slot alphabet and grammar of \citet{zattera2022}, the
crust--mantle--core grammar of \citet{stolfi2000}, and the standard EVA
composites (Appendix~\ref{app:inventory}).

\section{Results}

\subsection{A token is not yet a word: lexical appearance without prose-like order}
\label{sec:tokenorder}

The isolated ZL tokens make the word interpretation tempting. On static
vocabulary measures, Voynichese does not look obviously impoverished or
mechanical (Table~\ref{tab:tokenorder}). Its 7,022 distinct forms fall inside
the continuous-text range of 3,454--11,041. The 50 most frequent forms cover
33.2\% of tokens, also inside the control range of 24.0--50.3\%. Singletons
make up 69.7\% of the Voynich types, just above the control maximum of 67.7\%;
merging every uncertain separator raises this to 73.0\%. Thus the long tail is
unusually rich, but the overall number and concentration of forms remain
broadly word-like at this sample size.

\begin{table}[H]
\centering
\small
\caption{Static vocabulary and adjacent-token order. Continuous-text controls
match the Voynich token count and line-length template. The catalogue matches
token count while preserving record boundaries. Order values use a
2,000-type cap and subtract the mean of 100 within-line shuffles.}
\label{tab:tokenorder}
\begin{tabular}{lrrrrr}
\toprule
Corpus & Types & Singleton types & Top-50 coverage & Excess bits & Order share \\
\midrule
Voynich, observed spaces & 7,022 & 69.7\% & 33.2\% & 0.0661 & 0.79\% \\
Voynich, weak spaces merged & 7,920 & 73.0\% & 30.0\% & 0.0438 & 0.54\% \\
Latin narrative & 8,084 & 59.3\% & 26.7\% & 0.3256 & 4.00\% \\
English narrative & 4,248 & 51.7\% & 50.3\% & 0.7422 & 9.18\% \\
French narrative & 5,828 & 58.5\% & 44.5\% & 0.7828 & 9.72\% \\
German narrative & 6,920 & 63.4\% & 39.0\% & 0.3866 & 4.81\% \\
Italian narrative & 7,836 & 66.3\% & 37.2\% & 0.4810 & 6.18\% \\
Latin medical & 7,332 & 58.6\% & 32.0\% & 0.2927 & 3.59\% \\
Latin botanical & 11,041 & 67.7\% & 24.0\% & 0.1476 & 2.02\% \\
English herbal & 3,454 & 46.8\% & 47.0\% & 0.7947 & 9.55\% \\
\emph{Species Plantarum} records & 5,251 & 53.2\% & 38.5\% & 1.3478 & 15.81\% \\
\bottomrule
\end{tabular}
\end{table}

The contrast appears when sequence is restored. At the 2,000-type cap, the
observed Voynich order contributes 0.0661 bits beyond the shuffled baseline,
or 0.79\% of token entropy. This is measurable---the observed plug-in mutual
information exceeds all 100 shuffled values---but small. The lowest prose
control, botanical Latin, gives 2.02\%; the other continuous-text controls give
3.59--9.72\%. At this cap Voynichese is therefore below 40\% of even the
domain-matched minimum. The botanical catalogue reaches 15.81\%, almost twenty
times the Voynich share. Low order is not a generic consequence of botanical
subject matter or record-like genre.

Treating uncertain separators as possible false splits does not repair the
deficit: merging all 2,352 of them leaves 30,395 tokens and reduces the order
share from 0.79\% to 0.54\%. Nor does a single codicological section, the
Currier language, or the vocabulary cap drive the result (leave-one-quire-out
range 0.48--0.83\%; Currier A and B separately approximately zero; caps of
250--4,000 leave the Voynich share below the lowest control at every
resolution; Appendix~\ref{app:robust}).

\begin{figure}[tbp]
\centering
\includegraphics[width=\textwidth]{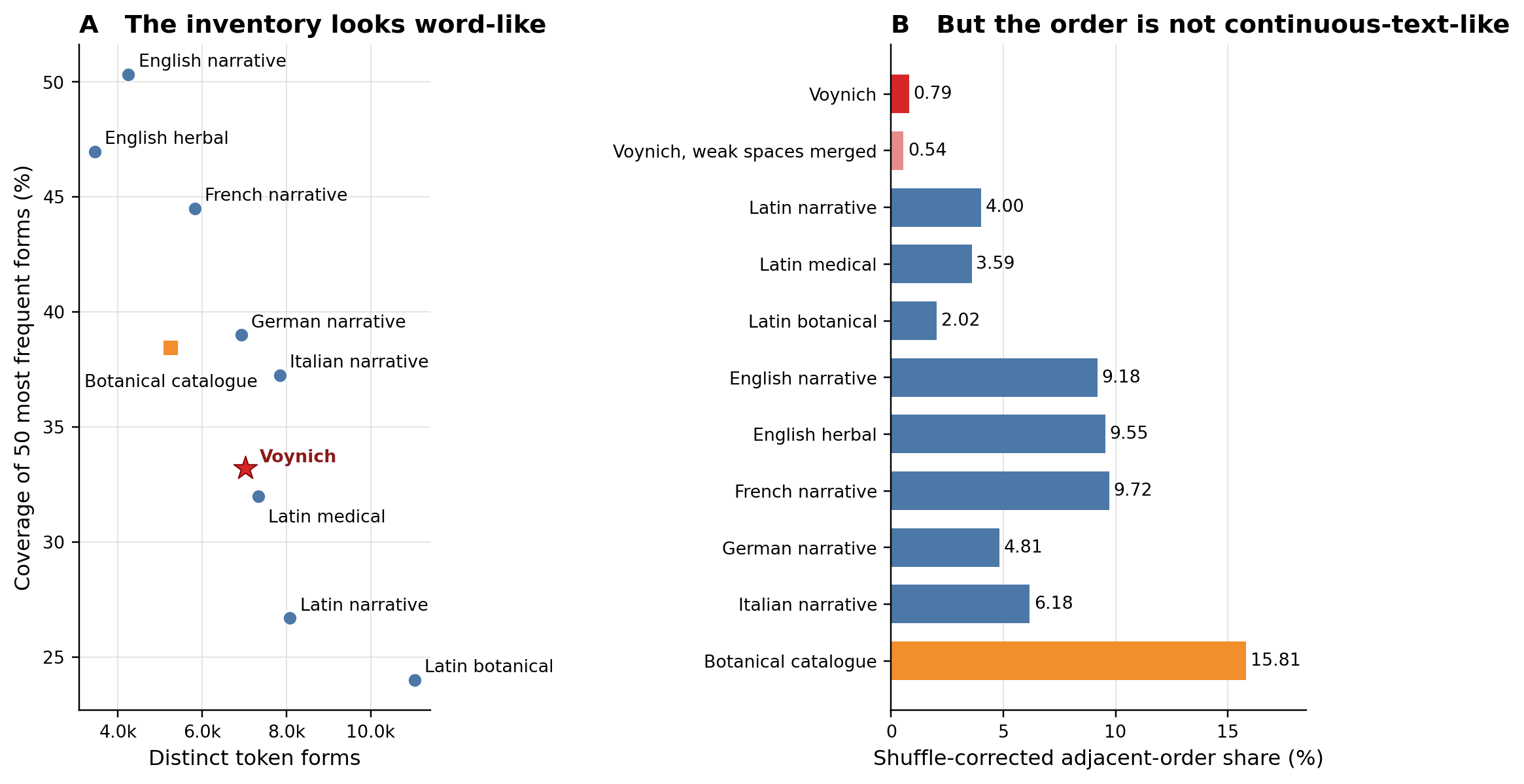}
\caption{A word-like inventory with little prose-like order. (A) Distinct
forms and coverage of the 50 most frequent forms at 32,747 tokens. Voynichese
falls among the controls on these static measures. (B) Adjacent-token mutual
information after subtracting the mean of 100 within-line shuffles, expressed
as a share of capped token entropy. Continuous-text controls span narrative,
medical, botanical, and herbal genres and match the Voynich line template;
the catalogue retains record boundaries. Values use a 2,000-type cap.}
\label{fig:scale-transition}
\end{figure}

The anomaly is therefore not that Voynichese lacks reusable forms. It is that
those forms predict the identity of their neighbours remarkably little. The
observed strings are valid transcription tokens, but their lexical appearance
is insufficient to identify them as the words of ordinary prose or catalogue
text. What order there is sits at the token edges, not in token identity
(Section~\ref{sec:edge}).

\subsection{Order lives at the token edges, not in token identity}
\label{sec:edge}

The token-identity result must be read together with a well-known fact
about Voynichese: the last glyph of one token constrains the first glyph of
the next \citep{currier1976,reddy2011,parisel2026a}. On the same corpora and
shuffle null, the shuffle-corrected information between adjacent edge glyphs
is 0.197 bits in Voynichese, 5.7\% of the entropy of the next token's first
glyph---close to the 0.23 bits reported by \citet{parisel2026a}---and larger
than in any continuous-text control (0.018--0.120 bits, 0.4--2.9\%); only the
Linnaean catalogue is higher (0.322 bits) (Table~\ref{tab:edge};
Appendix~\ref{app:edge}). Nearly half of it is a single pair, \(y\) followed by
\(q\), with \(r\) or \(s\) followed by \(a\) and the avoidance of \(y\) before
\(ch\) or \(sh\) supplying most of the remainder; it is present in both
Currier languages (A 0.137, B 0.248 bits) and is 88--93\% destroyed by
within-line shuffling, so it is a genuine ordering effect rather than a
line-composition one. The hierarchy is the reverse of every control: in the
languages, ordering information is carried by whole-word identity and leaks
only weakly into the boundary letters (edge-to-identity ratio 0.10--0.24),
whereas in Voynichese it is concentrated in the boundary glyph pair (ratio
2.9), widening the window to two glyphs adds only about a fifth (controls: a
factor of four to seven), and knowing the whole preceding token predicts the
next first glyph no better than its last glyph does. Adjacent Voynich tokens
are also within one edit of each other 12--16\% more often than under
shuffling, the effect emphasised by \citet{timm2020}, whereas every prose
control suppresses near-identical neighbours (ratios 0.28--0.56) and only the
catalogue amplifies them (2.6). Two cautions follow. The mixed statistic
``last glyph of \(t\) predicts the identity of \(t+1\)'' puts Voynichese at
2.0\% of token entropy, at the bottom of, rather than clearly below, the
continuous-text range (1.7--4.9\%), because the edge effect feeds it; and the
size of the identity deficit depends on the vocabulary cap (a factor of 0.8
below the lowest control at a 250-type cap, 0.4 at 2,000, and at the noise
floor at 4,000). What is robust is the ordering of the two scales. The
token-order deficit is a statement about \emph{where} adjacent order lives in
Voynichese---at the glyph edges, at the scale of the learned units---not
about the absence of sequential structure across separators.

\begin{table}[H]
\centering
\small
\caption{Where adjacent order lives. Shuffle-corrected mutual information (100
within-line shuffles) between adjacent tokens, at the edge-glyph scale and at
the token-identity scale (2,000-type cap), and the ratio of observed to
shuffled frequency of adjacent tokens within one edit of each other. Full
table in Appendix~\ref{app:edge}.}
\label{tab:edge}
\begin{tabular}{lrrrrr}
\toprule
& \multicolumn{2}{c}{Last glyph \(\to\) first glyph} &
\multicolumn{2}{c}{Token \(\to\) token} & Edit-1 \\
Corpus & bits & share & bits & share & ratio \\
\midrule
Voynich, observed spaces & 0.197 & 5.7\% & 0.067 & 0.81\% & 1.12 \\
Voynich, weak spaces merged & 0.162 & 4.8\% & 0.044 & 0.54\% & 1.14 \\
Voynich, Currier A & 0.137 & 3.9\% & \(-0.043\) & \(-0.50\%\) & 1.20 \\
Voynich, Currier B & 0.248 & 7.4\% & 0.029 & 0.35\% & 1.07 \\
Latin botanical (lowest prose) & 0.018 & 0.4\% & 0.148 & 2.03\% & 0.56 \\
French narrative (highest prose edge) & 0.120 & 2.9\% & 0.782 & 9.71\% & 0.53 \\
English herbal & 0.086 & 2.1\% & 0.794 & 9.54\% & 0.47 \\
\emph{Species Plantarum} records & 0.322 & 7.9\% & 1.348 & 15.81\% & 2.57 \\
\bottomrule
\end{tabular}
\end{table}

\subsection{A separator is not one uniform word boundary}
\label{sec:separator}

Weak token order would be uninformative if the token boundaries were simply
arbitrary transcription marks. They are not. Sequence statistics already
separate the two ZL blank classes before any image information is introduced.

\begin{table}[H]
\centering
\caption{Normalised internality by boundary class. Intervals directly resample
the 16 quires and recompute the full estimator. Rows are overlapping
cross-classifications, not one partition.}
\label{tab:boundary}
\begin{tabular}{lrrr}
\toprule
Boundary class & \(\Iidx\) & 95\% quire-bootstrap interval & \(n\) \\
\midrule
Within-token pair & 1.000 & anchor & -- \\
Uncertain separator \texttt{,} & 0.494 & [0.464, 0.524] & 2,409 \\
First separator of line & 0.099 & [0.052, 0.145] & 3,834 \\
Mid-line separator & 0.064 & [0.002, 0.146] & 21,836 \\
Certain separator \texttt{.} & 0.029 & [-0.032, 0.111] & 27,073 \\
Continuation line break & -0.036 & [-0.072, -0.002] & 3,634 \\
\bottomrule
\end{tabular}
\end{table}

Uncertain separators score \(\Iidx=0.494\), compared with \(0.029\) for
certain separators (Table~\ref{tab:boundary}). The difference of 0.465 index
units is much larger than the positional differences among ordinary
separators. First-of-line and mid-line estimates are slightly higher than the
certain-separator average, but their quire intervals overlap broadly. The line
break falls just below the independent-pair anchor.

The contrast is not concentrated in a section of the codex. On the common
full-corpus scale, mean uncertain-separator internality exceeds the certain
mean in all 16 quires; the quire-specific differences range from 0.309 to
0.529 (two-sided sign test \(p=3.05\times10^{-5}\)). Recomputing the complete
scale after omitting each quire in turn leaves an uncertain-minus-certain
contrast between 0.411 and 0.479. No single quire is necessary for the result.

\begin{figure}[tbp]
\centering
\includegraphics[width=0.90\textwidth]{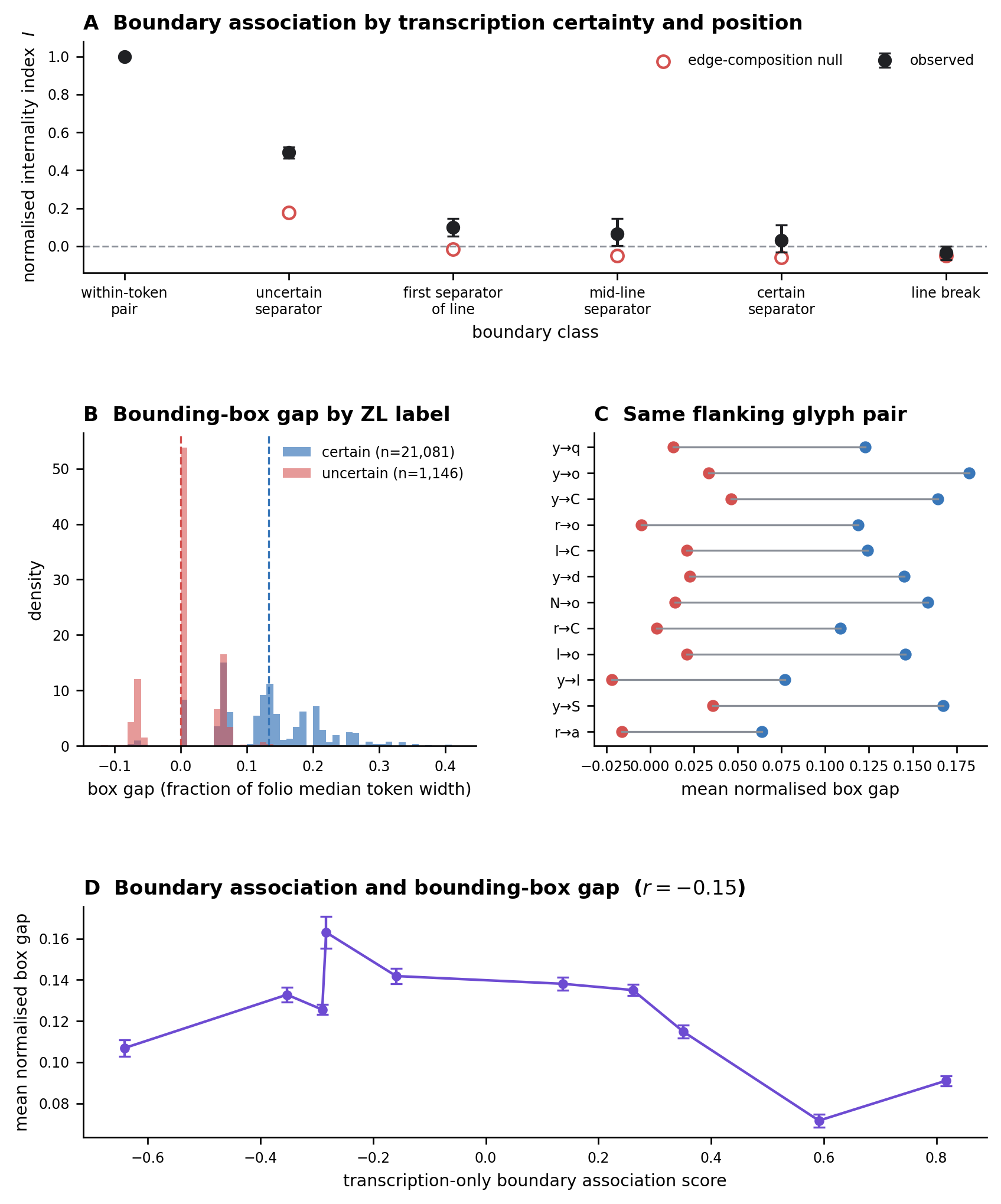}
\caption{Boundary structure by transcription and by image. (A) Boundary
association by transcription certainty and structural position; filled points
are observed values with quire-bootstrap intervals, open points are expectations
under the edge-composition null (not shown for the within-token anchor).
(B) Bounding-box gaps by ZL certain/uncertain label; dashed lines show medians.
(C) Mean gaps for the 12 most frequent flanking glyph pairs represented at both
labels; blue denotes certain and red uncertain, as in panel~B. (D) Mean box gap
across equal-count bins of the transcription-only boundary association score;
error bars are folio-bootstrap standard errors.}
\label{fig:boundary}
\end{figure}

The edge-composition null explains only part of the uncertain-separator score:
observed \(\Iidx=0.494\), null \(0.177\), residual \(0.317\). Residuals are
smaller for certain, first, and mid-line separators (0.090--0.116), and only
0.013 at line breaks. The uncertain category is therefore not produced solely
by different glyph frequencies at token edges.

\subsection{The separator regimes are physically recoverable}
\label{sec:coordinates}

Across the 223 coordinate files, 28,626 of 35,439 visual tokens (80.8\%) fall
in exact aligned blocks. Those blocks yield 22,227 labelled intra-line
separators on 204 folios: 21,081 certain and 1,146 uncertain.

The mean normalised box gap is 0.1285 at certain separators and 0.0043 at
uncertain separators, a difference of 0.1242. The folio-bootstrap 95\%
interval is [0.1194, 0.1288]. The median is 0.1333 for certain separators and
0.0000 for uncertain separators. The zero means that the two boxes meet or
overlap at the 3-pixel grid resolution; it is not a claim that the ink itself
has zero separation.

The direction is highly consistent across the manuscript. Among 179 folios
containing both labels, 176 have wider mean gaps at certain separators. More
importantly, controlling for the flanking glyph pair gives
\(\widehat{\beta}=-0.0991\), with folio-bootstrap interval
[-0.1036, -0.0946]. All 40 glyph pairs occurring at least five times with both
labels have a wider mean gap when labelled certain. Adding folio fixed effects
leaves the uncertain-label coefficient essentially unchanged at -0.0976.

The physical distinction is large enough to recover the transcription label,
not merely to shift a mean. Bounding-box gap alone gives AUC 0.9053
(folio-bootstrap 95\% interval [0.8972, 0.9127]). In leave-one-folio-out
validation, a threshold learned on the remaining folios identifies uncertain
separators with 86.7\% sensitivity and 80.1\% specificity, for balanced
accuracy 83.4\%. Because no held-out labels enter threshold selection, this
shows that the distinction generalises across folios.


The continuous boundary score and physical box gap are negatively associated
(\(r=-0.152\)). The relationship is not monotonic at every score bin, so we do
not interpret the score as a calibrated physical-gap predictor. Its value is
that a statistic derived from glyph sequences alone covaries with a physical
measurement from a source that did not contain the ZL certainty labels.

A stricter line-by-line alignment (17,737 separators) gives the same contrast
(0.1272 versus 0.0023). Nor is the distinction peculiar to one transcriber:
ZL and Claston's v101 agree on 93.1\% of 11,306 comparable separator labels
(Cohen's \(\kappa=0.310\), modest because uncertain marks are rare and the two
readers place them differently), and the physical result strengthens when the
labels are combined---mean normalised gap 0.1273 where neither marks
uncertain, 0.0300 where one does, \(-0.0044\) where both do
(Appendix~\ref{app:robust}). Independent uncertainty votes show a monotonic
dose response in image geometry.

\subsection{The two separator regimes are stable across Currier A and B}

The uncertain-separator effect is not confined to one dialect. Separators
flagged uncertain score \(I=0.494\) in Currier A and \(0.503\) in Currier B,
each stratum normalised against its own anchors (Appendix~\ref{app:dialect});
the difference of \(-0.009\) (A\,--\,B) has a 90\% quire-bootstrap interval of
\([-0.052,+0.037]\), which narrowly misses a pre-specified \(\pm0.05\)
equivalence margin while giving no evidence of a practically substantial
dialect difference. A confound check points the same way: if the effect merely
tracked how often a transcriber hesitated, the stratum with the higher
uncertain rate should show the larger raw gap, but Currier A has the higher
uncertain share (0.088 versus 0.080 of intra-line separators) and the smaller
raw gap between uncertain and certain mean scores (\(+1.53\) versus \(+2.52\)
bits).

The \emph{ordinary} boundaries, by contrast, differ between the dialects
(Figure~\ref{fig:dialectline}, left). Currier A's certain, mid-line, first-of-line, and
line-break separators all score above Currier B's (0.094 versus \(-0.003\);
0.125 versus 0.037; 0.104 versus 0.069; \(-0.010\) versus \(-0.041\)). What
produces this difference is left open: different languages, registers, topics,
encoding tables, or section conventions could each contribute, and scribal
hand is confounded with dialect \citep{fagindavis2020}. Because \(I\) is normalised
within each stratum against that stratum's own within-token cohesion, a
difference in the within-token anchor can move the normalised score even when
raw boundaries are identical. The A/B pattern is therefore descriptive, not a
mechanistic diagnosis.

\subsection{The line-start effect is a mixture of two line populations}

\begin{figure}[tbp]
\centering
\includegraphics[width=\textwidth]{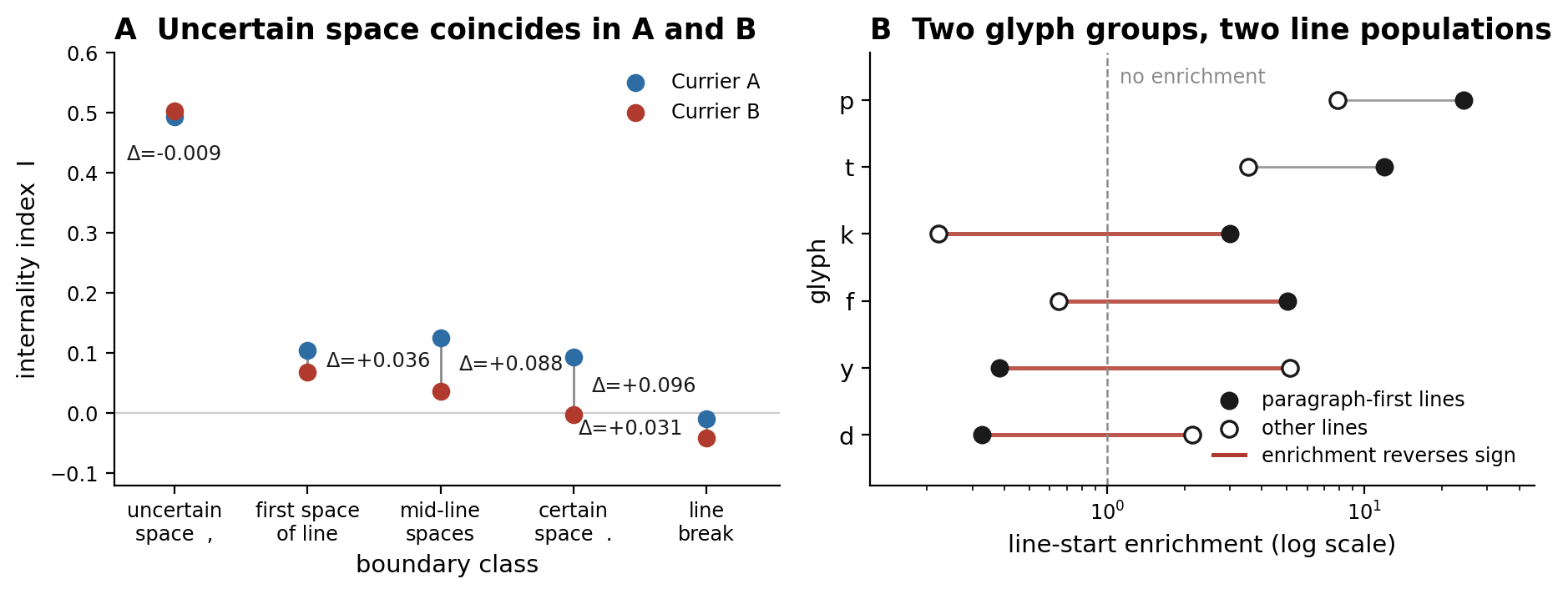}
\caption{(A) Normalised internality index computed separately for Currier A
and B. (B) Line-start enrichment of six glyphs, paragraph-first lines against
all other lines, relative to the mid-line word-initial rate (log scale;
connectors mark a sign reversal).}
\label{fig:dialectline}
\end{figure}

That the glyph opening a line differs sharply from the glyph opening an
ordinary mid-line token, and that gallows cluster in the first lines of
paragraphs, has been known since Currier and D'Imperio and has been quantified
several times \citep{currier1976,dimperio1978,vogt2012,feaster2021,steckley2024}.
What we add is a decomposition. Splitting lines by paragraph position gives a
line-initial Jensen--Shannon divergence (line-initial glyphs against mid-line
token-initial glyphs of the same lines) of 0.528 for the 717 paragraph-first
lines versus 0.179 for the 3,163 others; the pooled value of 0.203 conceals two
line populations differing by a factor of about 2.9 (quire-bootstrap 95\%
interval of the difference [0.31, 0.40]; Appendix~\ref{app:dialect}).
Paragraph-first lines open with gallows---\(p\) 24-fold, \(t\) 12-fold,
\(f\) 5-fold, and \(k\) 3-fold enriched relative to mid-line token-initial
rates---and almost never with \(y\) or \(d\); ordinary lines still favour
\(p\) and \(t\) but reverse the sign for \(k\), \(f\), \(y\), and \(d\). Lines
that open a page are the extreme case (JSD 0.49; \(k\) 6.4-fold). The two
glyph groups reported in earlier work are therefore signatures of two line
populations with different opening phases, not one homogeneous ``special
first-symbol'' rule (Figure~\ref{fig:dialectline}, right).

\subsection{A glyph is not licensed as a plaintext letter}

\begin{table}[H]
\centering
\caption{Marginal symbol entropy, first-order conditional entropy, and
conditional perplexity. All control corpora are truncated to the Voynich token
count. Composite-collapsed EVA is the primary glyph-like representation;
decomposed EVA is a sensitivity analysis.}
\label{tab:entropy}
\begin{tabular}{lrrr}
\toprule
Corpus & \(H(X)\) & \(H(X_{t+1}\mid X_t)\) & \(2^{H(X_{t+1}\mid X_t)}\) \\
\midrule
Voynichese, composite-collapsed EVA & 3.98 & 2.69 & 6.45 \\
Voynichese, decomposed EVA & 3.89 & 2.32 & 4.98 \\
Pliny, botanical Latin & 4.01 & 3.51 & 11.39 \\
Celsus and Caesar, Latin & 3.99 & 3.47 & 11.10 \\
Culpeper, English herbal & 4.16 & 3.52 & 11.50 \\
Italian prose & 4.01 & 3.45 & 10.91 \\
\bottomrule
\end{tabular}
\end{table}

Marginal entropy is similar across the corpora, but composite-collapsed
Voynich conditional entropy is 0.76--0.83 bits below the full-sample controls
(Table~\ref{tab:entropy}). Equivalently, the effective number of next-glyph
choices is 6.45, compared with 10.91--11.50, a reduction of 41--44\%.
Decomposing the standard EVA composites does not close the gap: it lowers
conditional entropy further, to 2.32 bits (4.98 effective choices).

Equal-length sampling does not explain the result. Across 300 contiguous
141,310-symbol windows from each control (1,200 windows total), the lowest
conditional entropy is 3.379 bits, still 0.689 bits above Voynichese. Nor is
the gap created by joining transcribed tokens: using only within-token
transitions gives 2.43 bits for Voynichese and 3.10--3.21 bits for the
controls.

A fixed one-to-one substitution cannot produce this difference from any of
the four controls because it changes symbol names, not their probabilities or
adjacent-pair structure. This is not a general exclusion of letter-based
encoding. Nulls, homophonic or verbose substitution, abbreviations,
context-dependent alphabets, positional variants, and strongly formulaic
source text can all change the measured conditional entropy. The result
excludes the conventional monoalphabetic-substitution null for sources with
conditional structure in the tested range, consistent with earlier reports of
unusually predictable Voynich glyph sequences
\citep{reddy2011,montemurro2013}.

The warranted conclusion is narrower than ``Voynichese has no letters'' but
stronger than a visual resemblance. Under either explicit EVA inventory, a
transcription symbol is not licensed as a one-to-one plaintext letter. Any such
identification requires a source or encoding model that independently explains
the observed conditional structure.

\subsection{Recurrent multi-symbol units provide the supported scale}
\label{sec:units}

On the decomposed EVA stream, adjacent dependence falls from \(D_0=1.595\) bits to a
minimum \(D_{64}=1.045\). At that point the representation contains 88 observed
unit types and 72,512 unit occurrences, with mean span 2.39 symbols. Dependence
then rises to 1.119 bits at 128 merges, 1.475 at 256, and 3.044 at 1,024. The
early frequent learned units include \texttt{ol}, \texttt{ok}, \texttt{ar},
\texttt{ot}, \texttt{aiin}, \texttt{ch}, and \texttt{qok}; the algorithm is
recovering recurring structure rather than only producing rare long strings.

The trough is not a pooled-corpus accident. Currier A and B both attain their
minimum at 64 merges, with gaps 1.058 and 1.137 bits. Every one of the 16
leave-one-quire-out refits also selects 64 from the tested checkpoints. Across
200 random half-quire fits, 137 select 64 and the other 63 select 32; no fit
selects a smaller or larger checkpoint. At 64 merges the compound-unit sets
learned from disjoint quire halves overlap by a median 83.9\% of the smaller
set (central 95\% of partitions: 74.2--88.9\%).

The more demanding cross-fit moves the exact minimum but preserves the early
scale transition. When merge rules are learned on 15 quires and scored only on
the omitted quire, the glyph-weighted dependence gaps are 1.686, 1.423, 1.379,
and 1.490 bits at 0, 16, 32, and 64 merges. The held-out minimum is therefore
32, followed by a rise at 64. This rules out the interpretation that the trough
is merely the in-sample reward for fitting a segmentation to the same word
types. The defensible result is a reproducible early multi-symbol regime around
32--64 merges, not a privileged universal inventory of exactly 88 units. The
units it contains are not arbitrary: every compound unit learned by 64 merges
is a legal string of Zattera's 12-slot word template and 98\% of their
occurrences are paths through his grammar, yet only five of them are his
slot characters, and a third of all occurrences are fragments that cross the
layer boundaries of Stolfi's crust--mantle--core grammar
(Appendix~\ref{app:inventory}). The compression scale that minimises adjacent
dependence lies between the slot character and the token---at the
circle-modified letter (\texttt{ol}, \texttt{ok}, \texttt{ar}) and the
multi-slot suffix (\texttt{edy}, \texttt{aiin})---which is where the paper's
``recurrent multi-symbol unit'' should be understood to sit.

\begin{figure}[tbp]
\centering
\includegraphics[width=\textwidth]{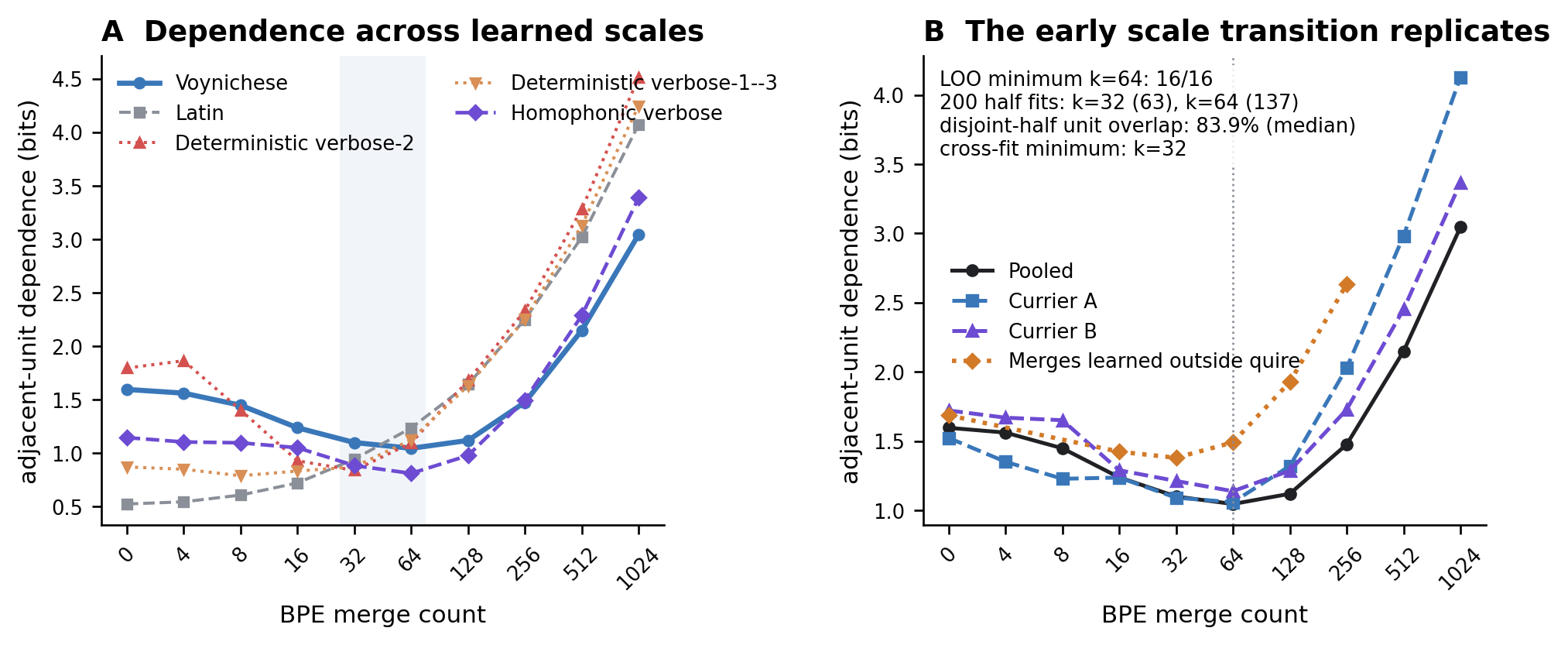}
\caption{Scale-dependent adjacent-unit structure under within-token BPE.
(A) Dependence gap across BPE merge counts for Voynichese, Latin, two
deterministic verbose substitutions, and a homophonic verbose substitution.
All streams contain approximately 173,000 input symbols. The shaded band marks
the 32--64-merge trough. (B) Currier A, Currier B, and the pooled corpus all
reproduce a minimum at 64 merges. The orange dotted curve learns its merges
outside each omitted quire and scores the held-out quire only; it reaches its
minimum at 32. Stability annotations summarise quire-level refits; overlap is
intersection size divided by the smaller learned-unit set in each disjoint-half
partition.}
\label{fig:unitscale}
\end{figure}

The controls clarify what this result does and does not mean. Plain Latin has
its lowest gap before merging and rises from 0.524 to 1.235 bits by 64 merges.
The verbose controls show troughs, with \(D_{64}=1.095\) and 1.124 bits for the
two deterministic encodings and 0.811 for the homophonic encoding. Voynichese
therefore has the qualitative scale signature expected when fine-grained symbols combine
into recurrent larger units, but the tested curves do not uniquely distinguish
verbose from homophonic or non-cryptographic chunking.

\begin{figure}[tbp]
\centering
\includegraphics[width=\textwidth]{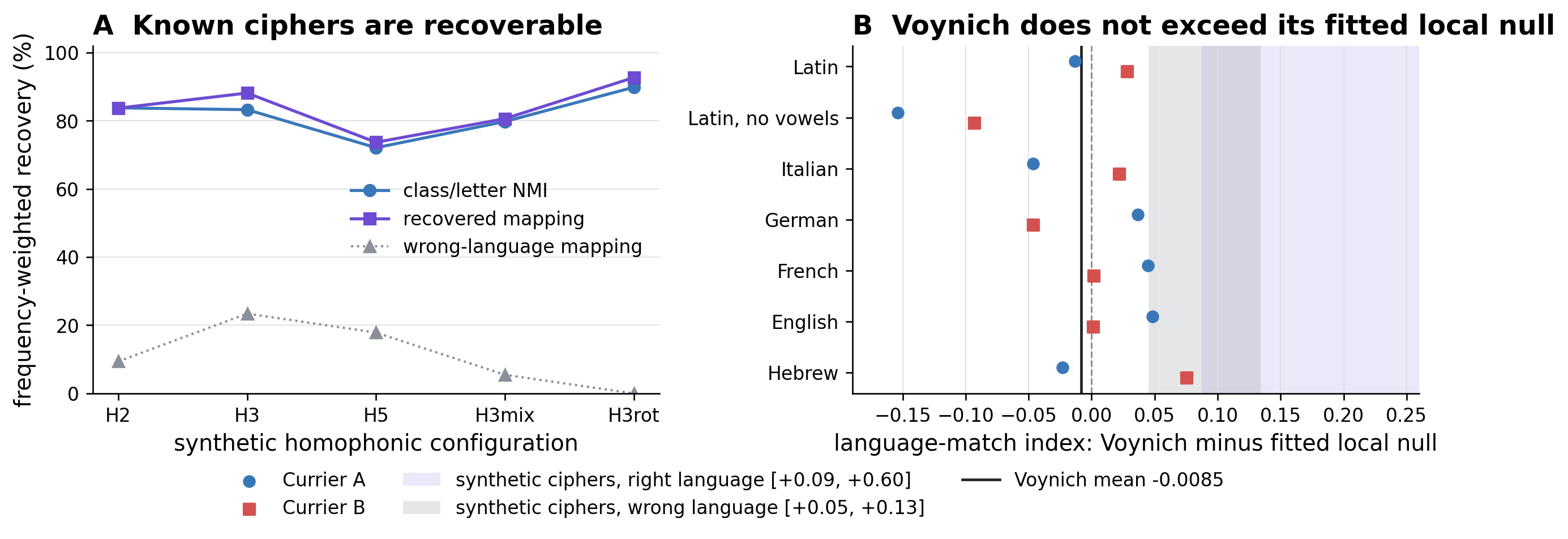}
\caption{Calibration and application of the learned-unit substitution attack.
(A) Ground-truth class/letter association and attacked mapping recovery across
five synthetic homophonic configurations of the same Latin; the wrong-language
mapping is a contrast. (B) Every principal Voynich-minus-fitted-local-null
language-match differential for Currier A and B; the solid vertical line is
their mean. Shaded bands are the range of the same differential for the
synthetic ciphers over five seeds each, against the right language (violet)
and against a wrong language (grey).}
\label{fig:cipher}
\end{figure}

That local compatibility does not survive the separately calibrated
decipherment test. Across the five synthetic configurations the pipeline
recovers 73.7--92.6\% of the frequency-weighted mapping, against 0--23.4\% for
the wrong-language mapping (Figure~\ref{fig:cipher}A). Over five cipher seeds
per configuration, the real-minus-surrogate differential against the right
language is positive in all 25 runs, with configuration means from \(+0.13\)
(variable one-to-three groups) to \(+0.52\) (rotating tables) and \(+0.33\)
for the three-homophone cipher; against a wrong-language model the same
ciphers give \(+0.05\) to \(+0.13\) (Appendix~\ref{app:controls}). Across the
14 principal Currier-by-language-model comparisons, the Voynich-minus-surrogate
differential ranges from \(-0.154\) to \(+0.075\) and averages \(-0.0085\)
(Figure~\ref{fig:cipher}B): below the right-language band of every synthetic
cipher, and at or below the band that a real cipher shows against the
\emph{wrong} language. The fitted short-range generator reproduces the
apparent language match of Voynichese for every tested model.

This rejects the calibrated family of simple concatenative homophonic prose
ciphers at the one-to-three-symbol scale. The BPE curves alone do not exclude
verbose or homophonic encoding, and deterministic verbose substitution was not
a separate end-to-end calibration. Null insertion was not tested at all.
Transposition, null systems, larger groups, atypical plaintexts, and
state-dependent encodings therefore remain outside this result---and
Section~\ref{sec:controls} shows concretely that a published verbose
homophonic cipher with respacing falls outside it.

The scale change also reframes the low-entropy observation. Raw adjacent
dependence is 41.3\% of Voynich marginal entropy, versus 13.3\% in Latin. At
64 merges, where the inventories are nearly matched (88 and 86 types), the
shares are 17.6\% and 20.6\%, respectively. Thus the exceptional dependence of
decomposed Voynich symbols does not survive the learned-unit representation. This is a
positive unit-scale result, not evidence that sequential structure disappears.

\subsection{Two accounts that earn part of the profile}
\label{sec:controls}

The joint profile is only useful if it discriminates, so we ran two published
accounts of Voynichese through the same drivers (Table~\ref{tab:profile};
Appendix~\ref{app:controls}). The Naibbe cipher of \citet{greshko2025} is a
verbose homophonic substitution built to imitate the manuscript: it deletes
plaintext spaces, respaces the letters into singles and pairs, and renders
each unit as one of six table-specific EVA strings, so a ciphertext token
encodes one or two plaintext letters. Encrypting the Latin control with our
re-implementation of the published tables (and, as a check, Greshko's own
Pliny ciphertext) reproduces the manuscript's glyph-level profile almost
exactly: conditional entropy 2.71/2.34 bits, a BPE trough at 64 merges of
1.03--1.04 bits with 82 units of span 2.4, learned units that cross only
1.9--2.2\% of the hidden token boundaries after space erasure, and a token
order share of 1.1--1.3\%. It also defeats the calibrated attack: the
language-match differential against the held-out Latin model is
\(+0.002\) (Caesar) and \(-0.008\) (Pliny) over five seeds, inside the Voynich
range and far from the \(+0.13\) to \(+0.52\) of the concatenative synthetic
ciphers, because a Naibbe letter is spread over about 3.4 glyphs and eighteen
homophones, outside the one-to-three-symbol family the attack was calibrated
on. The self-citation generator of \citet{timm2020}, ported from their
published rules and calibrated only to token length and vocabulary size,
likewise reproduces the low conditional entropy (2.57--2.70 bits), a trough at
64 merges in every seed, a token order share of 0.5--0.6\%, an excess of
near-identical neighbours of the same size as the manuscript's, and a null
attack differential (\(-0.02\) to \(+0.02\)).

Neither account earns the rest. Both have almost no edge-glyph information
(0.006--0.009 bits, 0.2--0.3\% of the next first glyph's entropy, against
0.197 bits, 5.7\%, in Voynichese) and neither reaches the manuscript's
hapax-rich vocabulary (41--42\% and 59--60\% singleton types against 70\%). The
self-citation streams additionally have a much deeper trough (0.65--0.86
bits) and vary widely between seeds (5.0--7.1 thousand types), whereas the
Voynich values are stable across quires and Currier languages; the Naibbe
streams have a more closed vocabulary (5.2--5.5 thousand types) and slightly
more token order than the manuscript. Neither has a second class of
separator, so the graded-space regime cannot be tested on them. Four members
of the profile---low glyph entropy, the learned-unit scale, weak token
identity order, and the null attack differential---are therefore
reproducible by both a Voynich-imitating cipher and a meaningless generator,
and do not by themselves discriminate; the edge-glyph coupling, the vocabulary
tail, and the graded boundaries are where an account of the text has still to
earn its claim.

\begin{table}[H]
\centering
\scriptsize
\setlength{\tabcolsep}{2.6pt}
\caption{The profile applied to two published accounts. \(h_2\): first-order
conditional entropy of composite-collapsed glyphs. \(D_{64}\): BPE dependence
gap at 64 merges (\(D_0\) is 1.56--1.63 for all Voynich-like rows and 0.52 for
Latin). Order: shuffle-corrected token-identity share at the 2,000-type cap.
Hapax: singleton share of types at 32,747 tokens. Edge: shuffle-corrected
last-glyph\(\to\)first-glyph information. Edit-1: observed/shuffled ratio of
adjacent tokens within one edit. Differential: language-match index minus that
of the order-3 surrogate against the held-out Latin model (Voynich: range over
14 model comparisons; Naibbe: mean of five seeds; self-citation: range over
three seeds). Crossing: share of hidden token boundaries crossed by learned
units after space erasure (Voynich: certain separators). Details in
Appendix~\ref{app:controls}.}
\label{tab:profile}
\begin{tabular}{lrrrrrrrr}
\toprule
Stream & \(h_2\) & \(D_{64}\) & Order & Hapax & Edge & Edit-1 & Differential & Crossing \\
\midrule
Voynichese & 2.69 & 1.05 & 0.79\% & 70\% & 0.197 & 1.12 & \(-0.15\) to \(+0.08\) & 2.5\% \\
Naibbe cipher of Latin (Caesar) & 2.71 & 1.03 & 1.25\% & 41\% & 0.006 & 0.82 & \(+0.00\) & 2.2\% \\
Naibbe cipher of Latin (Pliny) & 2.71 & 1.04 & 1.06\% & 42\% & -- & -- & \(-0.01\) & 1.9\% \\
Self-citation generator (3 seeds) & 2.6--2.7 & 0.65--0.86 & 0.5--0.6\% & 59--60\% & 0.006--0.009 & 1.06--1.23 & \(-0.02\) to \(+0.02\) & -- \\
Latin narrative & 3.44 & 1.24 & 4.00\% & 59\% & 0.033 & 0.43 & (index 0.99) & 15.5\% \\
Synthetic homophonic ciphers & -- & 0.8--1.1 & -- & -- & -- & -- & \(+0.13\) to \(+0.52\) & 3.8--12.4\% \\
\bottomrule
\end{tabular}
\end{table}

\subsection{A token is not a syllable}
\label{sec:syllable}

The letter and word readings are not the only unit hypotheses. Stolfi's
long-standing proposal that each token is one syllable of a tonal, isolating
language of the East Asian type, written in an invented phonetic script
\citep{stolfi1997chinese,stolfi2002}, makes token-level predictions that can be checked against
pinyin renderings of a genre-matched classical Chinese herbal (\emph{Bencao
Beiyao}, 1694) and a narrative (\emph{Romance of the Three Kingdoms}) at
matched token counts (Appendix~\ref{app:syllable}, Table~\ref{tab:syllable}).
Two comparisons carry the weight. Shuffle-corrected adjacent order is
5.2--6.4\% of capped entropy for the pinyin syllable streams, including the
herbal, against \(-0.5\%\) and \(+0.4\%\) for Currier A and B and 0.79\% pooled;
and at a matched sample of about 10,700 tokens the pinyin streams use 335--749
syllable types with 10--20\% hapax, whereas Currier A uses 3,343 types with
72\% hapax. A syllabary is a closed inventory whose syllables carry sequential
structure; Voynich tokens are neither closed nor sequentially constrained. Two
further checks---the merge-scale profile of the pinyin letter stream and the
calibrated attack against a pinyin-letter trigram model---point the same way
but are less specific. \citet{reddy2011} already noted that Voynich letter
statistics resemble pinyin; the order and closure comparisons show that the
resemblance stops at the token level. The test covers Mandarin only; languages
with larger syllabaries would narrow the inventory gap but not the order gap.

\subsection{Erasing spaces preserves the learned scale and boundary hierarchy}

Space erasure leaves the BPE curve almost unchanged. On the strict-separator
subset, the observed-token curve has \(D_{32}=1.102\), \(D_{64}=1.049\),
\(D_{128}=1.139\), and \(D_{256}=1.500\) bits. Fitting the continuous line
strings instead gives 1.115, 1.072, 1.181, and 1.581 bits, respectively, and
retains the minimum at 64 merges. The unit-scale result therefore does not
depend on forbidding merges across the transcribed spaces.

Arbitrary constraints fit worse. With uniformly random separator positions,
the median \(D_{64}\) is 1.241 bits (central 95\% of 50 fits: 1.230--1.254);
41 fits select 64 merges and nine select 32. When the observed token-length
multiset is preserved within every line but reassigned to different glyph
positions, all 50 fits select 32 merges and median \(D_{64}\) is 1.271 bits
(1.256--1.280). Thus neither separator count nor the token-length distribution
explains the observed 64-merge curve.

The erased-space segmentation also recovers the separator hierarchy without
seeing its labels (Table~\ref{tab:spacecross}). At 64 merges, learned units cross
only 2.5\% of ZL-certain separator locations, compared with 20.3\% of uncertain
locations and 60.0\% of all possible adjacent-glyph positions. Uncertain
crossing exceeds certain crossing in all 16 quires. The control rates show that
this diagnostic can recover genuine word boundaries, but do not make lexical
status an assumption or a conclusion.

\begin{table}[H]
\centering
\caption{Crossing of hidden separator positions by units learned from
space-erased line strings at 64 BPE merges. Control separators are genuine
Latin plaintext word boundaries retained through encoding.}
\label{tab:spacecross}
\begin{tabular}{lrrr}
\toprule
Hidden position class & Crossed & Total & Crossing rate \\
\midrule
Voynich certain separator & 661 & 26,447 & 2.5\% \\
Voynich uncertain separator & 477 & 2,350 & 20.3\% \\
Voynich all intra-line positions & 97,402 & 162,330 & 60.0\% \\
Latin word separator & 3,796 & 24,464 & 15.5\% \\
Deterministic verbose-2 word separator & 1,528 & 12,360 & 12.4\% \\
Homophonic verbose word separator & 464 & 12,360 & 3.8\% \\
\bottomrule
\end{tabular}
\end{table}

The raw flanking pair alone also predicts space presence out of quire:
held-out AUC is 0.980 and the reduction from marginal cross-entropy is 71.9\%
over 162,330 positions, positive in every quire fold. This high predictability
rejects random placement, but it is compatible with either linguistic
morphotactics or a rule-governed surface process.

\section{Robustness and limitations}

\subsection{What weak token order does and does not show}

The token-order comparison is deliberately local: it measures adjacent forms,
not paragraph-wide syntax or semantic coherence. A text could have weak
bigram order yet carry information in long-range dependencies, page layout,
illustration links, numerical structure, or an untested segmentation. Nor do
eight continuous-text controls and one catalogue define all natural language.
They do, however, include domain-matched botanical, medical, and herbal prose
as well as five narrative-language conditions. The result should therefore be
read as a constraint on these continuous and record-like texts at the tested
units, not as a universal language test.

Three design choices reduce narrower artefacts. The shuffle correction removes
the large positive plug-in bias caused by rare token pairs. The continuous-text controls
have exactly the same token count and line-length template. And the separation
holds at every vocabulary cap and after every quire omission. Two limits of
the estimator should nevertheless be stated. At the 2,000-type cap the
observed Voynich plug-in value is 98\% shuffle bias, so the corrected estimate
is small and noisy, and at the largest cap it can be slightly negative---the
noise floor, not negative mutual information. And the size of the deficit
relative to the lowest control shrinks at coarser caps (Appendix~\ref{app:robust});
the robust findings are that Voynichese is below every control at every cap
and that its adjacent order is concentrated at the token edges rather than in
token identity (Section~\ref{sec:edge}), the reverse of every control. Earlier
reports of word-adjacency structure compatible with language
\citep{amancio2013,montemurro2013} and of strong cross-boundary glyph
information \citep{parisel2026a} are therefore not contradicted; they measure
the edge scale, or longer-range co-occurrence, rather than adjacent token
identity.

Non-lexical spaces remain a serious alternative, but not an unrestricted one.
Merging all uncertain separators lowers the order estimate, while learning
units after erasing every space recovers the same 64-merge trough and leaves
97.5\% of certain positions uncrossed. A proposed hidden segmentation may
still differ from the manuscript blanks, but an account that routinely joins
across the strong-space class must explain why sequence learning and image
geometry both recover that class. Conversely, our tests do not prove that a
preserved strong space is a lexical word boundary.

\subsection{What the coordinate analysis does and does not establish}

The coordinate analysis validates a property of the transcription categories:
ZL-uncertain separators correspond to physically compressed box junctures.
The high AUC establishes recoverability of that palaeographic distinction
across folios; it does not reveal lexical status. A narrow gap may reflect
cursive practice, abbreviation, correction, or segmentation ambiguity rather
than a hidden within-word transition.

The coordinate provenance is incompletely documented. Variable box heights and
only partial correlation between box width and token length are consistent
with boxes drawn over the page images, but the coordinates are quantised to a
3-pixel grid, and no published construction protocol is available. The boxes
are, in effect, a third human reading of each juncture: whoever drew them
decided where one token ends and the next begins, and that decision is the same
kind of judgement a transcriber makes when placing a comma. The AUC therefore
measures agreement between two readings of the page as much as physical
geometry; the direct-pixel audit (Appendix~\ref{app:pixel}) shows a much weaker discrimination
on the ink itself. In addition, only junctures split by both tokenisation
systems can enter the analysis. Excluded joined cases may be especially
compressed, but the direction and magnitude of that selection effect are not
identified from the observed sample.

A related caveat concerns the association index itself. A transcriber is more
likely to mark a separator uncertain when the flanking glyphs would form a
familiar string if joined, so the ZL label partly embeds the within-token
association it is compared with. The edge-composition null removes marginal
edge preferences but not this pairing bias. What rescues the result is not the
index alone but its convergence with the box gaps, the ink gaps, and the
space-erased unit learning, in none of which the ZL comma enters the
measurement; it serves only as the stratifying label.

The v101 comparison reduces dependence on a single reader's judgement: the
0/1/2-vote gap gradient is stronger evidence than raw agreement alone. It is
still not a fully independent image experiment. Both transcriptions derive
from MS 408, only equal-separator-count lines enter the comparison, and v101's
glyph inventory differs from EVA. The check supports a reproducible weak-space
distinction; it does not turn either transcription into a lexical analysis.

\subsection{Scope of the learned-unit inference}

BPE is a frequency-based segmentation, not a linguistic oracle
\citep{sennrich2016}. Its units may
be ligatures, motor chunks, affixes, cipher groups, or recurrent fragments of a
generative template. At high merge counts every corpus accumulates rare longer
units, so the evidential feature is the stable early trough and its replication,
not the eventual increase in dependence.

The held-out-quire curve is the strongest guard against in-sample BPE
overfitting. Its minimum at 32 rather than 64 also prevents overinterpretation:
the data support an early multi-symbol scale, while the exact merge count and
the 88-unit full-corpus inventory remain analysis-dependent.

The result is not merely BPE undoing multi-character EVA conventions. After
the nine standard composites are collapsed first, the pooled gap is 1.311 bits
at \(k=0\), 1.049 at \(k=32\), 1.046 at \(k=64\), and 1.163 at \(k=128\).
The broad 32--64-merge trough remains. The primary BPE is constrained not to
cross transcribed spaces, but the separator-erasure fit removes that constraint
and retains the trough. Its crossing rates support a graded model: certain
spaces are strong distributional boundaries, while uncertain spaces are more
often permeable to learned units. This does not establish that either class is
a lexical word boundary or that the resulting BPE inventory is final.

Finally, the 16 quires are few top-level clusters, and eight continuous-text
controls do not exhaust linguistic typology. We therefore emphasise effect
sizes, direct cluster resampling,
leave-one-cluster stability, matched-length controls, and the exact scope of
the substitution invariance rather than universal language-level claims.

\section{Discussion}

\subsection{Observed units are not inferred linguistic units}

The manuscript gives us observable glyphs, tokens, and separators. Letters,
words, and word boundaries are hypotheses about those observations, and the
results reject a direct identification at each level while supplying a more
precise positive description. A transcription symbol is an observed coding
unit whose first-order conditional entropy is too low, under either EVA
inventory, for a fixed one-to-one relabeling of any tested alphabetic source;
the positive replacement is a reproducible scale of recurrent multi-symbol
units that absorbs most of the exceptional local dependence and survives
Currier splits, quire resamples, composite collapse, space erasure, and
out-of-quire application. Whether those units are linguistic, cryptographic,
palaeographic, or motor is not decided here. A ZL token is a blank-delimited
form with a plausible vocabulary size and common-form concentration---which
explains its resemblance to a word---but the preceding token predicts its
identity less than in any matched control at any vocabulary resolution; the forms
are well regulated internally, and their edges constrain one another, but
their identities are not ordered like the words of the tested prose or
records. A separator is a real structural event but not one class of word
space: certain and uncertain separators occupy different positions on the
transcription-only association scale, correspond to different physical gaps,
and remain differently permeable when every space is deleted before unit
learning. Neither class is thereby proven lexical. The methodological
consequence is direct: unit names must follow validation, not precede it.

\subsection{Implications for cipher hypotheses}

The joint result rules out more than either half alone. A fixed one-to-one
letter substitution cannot lower conditional entropy, so it cannot explain
the symbol-level result for sources in the tested range. A fixed one-to-one
codebook substitution of whole words likewise renames token identities without
removing their ordering information; without transposition, null insertion,
state dependence, or a different underlying segmentation, it cannot turn a
2--10\% continuous-text order share into 0.79\%. Figure~\ref{fig:unitscale}
by itself leaves verbose and homophonic encodings open because both controls
produce BPE troughs; Figure~\ref{fig:cipher} resolves part of that ambiguity,
since genuine concatenative homophonic ciphers of one to three symbols per
letter retain a language-over-surrogate differential of \(+0.13\) to
\(+0.52\), and \(+0.05\) to \(+0.13\) even against the wrong language,
whereas Voynichese averages \(-0.01\) and never exceeds \(+0.08\).

The Naibbe control shows exactly where that rejection stops. A verbose
homophonic cipher that respaces the plaintext and spreads each letter over
several glyphs and many homophones reproduces the glyph entropy, the learned
scale, the boundary permeability, and---within a factor of 1.5---the token
order of the manuscript, and its Latin is invisible to the attack. The
attack therefore rejects the calibrated concatenative family, not verbose
homophonic substitution with respacing, and a fortiori not transposition,
null systems, or state-dependent encodings. What such a cipher must still
earn is the strong coupling between the last glyph of one token and the first
of the next (0.2 bits in the manuscript, essentially none in Naibbe text,
whose token boundaries fall between unrelated plaintext chunks) and the
open, hapax-rich vocabulary (Naibbe: 41\% singleton types against 70\%). Both
are natural targets for a next cipher design; both are also exactly the
properties a decipherer should check before reading a proposed plaintext.

\subsection{Implications for language and pseudo-text hypotheses}

The token-order result is stronger than saying that Voynichese has an
unusual vocabulary, because its static vocabulary is not grossly unusual;
what separates it is the way those forms are sequenced, which explains why
word-frequency plots can look persuasive while word-by-word readings fail
\citep{landini2001,bowern2021}. Weak adjacent identity order is not,
however, equivalent to absence of meaning: lists, labels, paradigms,
litanies, tables, and formulaic notations can carry content without behaving
like continuous prose, and natural language under a different segmentation
remains possible, although frequent crossing of the strong spaces would
conflict with the space-erased unit analysis. The self-citation control cuts
the other way. A generator with no content, implemented from its authors'
own rules, earns the low entropy, the trough, the weak order, the excess of
near-identical neighbours, and the null differential; those features are
therefore consistent with meaningless text but do not establish it, and the
same features are consistent with the Naibbe cipher of a real text. What the
generator does not earn---the vocabulary tail, the edge-glyph coupling, a
trough of the right depth, and stability across sections---is where a test of
the hoax hypothesis would have to bite, and where, as far as we can measure,
the manuscript remains unexplained by either published mechanism.

\section{Conclusion}

The main conclusion is a change in both terminology and burden of proof. An
EVA glyph should be called a glyph until a model demonstrates that it functions
as a letter. A ZL token should be called a token or surface form until a model
demonstrates that it functions as a word. A transcribed blank should be called
a separator until its boundary function is established. Treating the three
observed categories as letters, words, and uniform word spaces builds an
untested decipherment into the data.

The evidence provides positive alternatives. Fine-grained symbol regularity
resolves onto a quire-stable early scale of recurrent multi-symbol units, with
an out-of-quire minimum at 32 merges and a full-corpus minimum at 64.
Blank-delimited forms constitute a plausible static inventory but carry less
adjacent identity information than any continuous-text control at any
vocabulary cap, while their edge glyphs carry more than any prose control.
Separators form strong and weak regimes that can be recovered from sequence,
image geometry, and learning after every space has been erased. Voynichese is
therefore not structureless; its structure is concentrated within forms and
at graded boundaries rather than in prose-like token succession.

This profile, rather than any single entropy or frequency statistic, is the
paper's deliverable, and we have measured how much of it two published
accounts already earn. A Voynich-imitating verbose cipher and a self-citation
generator both reproduce the 2.7-bit conditional entropy, the learned-unit
scale, the below-1\% token-order share, and the null cipher differential;
those four features are therefore necessary but not discriminating. Neither
reproduces the 0.2-bit edge-glyph coupling, the 70\% singleton vocabulary,
or---untestable on them---the two separator regimes (uncertain-separator
internality 0.494 against 0.029, corroborated by independent layout
coordinates at AUC 0.905 and by a blind ink-gap measurement). A model that
reproduces only word lengths, Zipf-like frequencies \citep{zipf1949}, or the
visual appearance of text has explained why Voynichese looks familiar; a model
that also reproduces the entropy, the units, and the weak order has explained
why it resists the standard attacks. Neither has yet established letters,
words, or word spaces, and the remaining discriminators are now named.

\section{Data and code availability}

The ZL transliteration, control corpora, per-token coordinate files, and
public analysis archive are available at
\url{https://github.com/lrozanova/voynich-units} (analysed snapshot: commit
\texttt{66f8adaadc120f93e8a4c906685ba66a9d3ab847}); source provenance is
documented there and in the cited records \citep{beinecke,zl3b,voynichese,
perseus,coffee2013,gutenbergcontrols}. The publication package contains one
driver per analysis---\path{reproduce_headlines.py} (boundary index, entropy,
coordinates, v101 check), \path{reproduce_scale_transition.py} (token order),
\path{reproduce_edge_order.py} (edge-glyph order),
\path{reproduce_unit_scale.py} (learned units),
\path{reproduce_unit_inventory.py} (comparison with published inventories),
\path{reproduce_space_sensitivity.py} (separator erasure),
\path{reproduce_dialect_linestart.py} (Currier strata and line start),
\path{reproduce_cipher_calibration.py} with the bundle's
\path{validate_synthetics.py}, \path{attack_voynich.py}, and
\path{run_differentials.py} (calibrated attack),
\path{reproduce_naibbe_control.py} and
\path{reproduce_selfcitation_control.py} (the two further controls), and
\path{syllabic_test.py} (syllabic hypothesis)---together with the data,
control corpora, and archived fixed-seed outputs they read, including the
transcribed Naibbe tables and Greshko's sample ciphertext (redistributed under
their modified MIT licence) and the generated self-citation streams under
\path{data/controls}. Every driver
fixes its seeds and takes its input roots as explicit arguments. The
Beinecke page images used by the direct-pixel audit are not redistributed.
The same package accompanies the arXiv version of this paper as ancillary
files.

\section*{Acknowledgements}

Large language models (Claude Fable 5, Anthropic; GPT~5.6 Sol, OpenAI) were
used to assist with data analysis code and with editing the manuscript. All
analyses, results, and interpretations were designed, checked, and approved by
the authors, who take full responsibility for the content.

\appendix
\small
\section{Estimator and reproducibility details}

\subsection{Boundary-index specification}

The collapsed boundary-corpus vocabulary contains 34 observed symbols. Add-0.5
smoothing is applied to every cell of the \(34\times34\) within-token bigram
table before normalisation. The same position-independent unigram vector supplies the two
PMI marginals and the independent-product expectation. Bootstrap duplicates of
a quire duplicate its lines and counts before all tables and anchors are
recomputed.

The five reported boundary pools are constructed as follows: uncertain and
certain use the ZL comma/dot label; first and mid use separator position within
the line; and continuation line breaks join consecutive non-paragraph-first
lines on the same page. Last-of-line intra-line separators are included in the
certain/uncertain totals but not shown as a separate main-table row.

\subsection{Coordinate fixed effect}

The pair-fixed-effect coefficient is estimated after demeaning both box gap and
the uncertain-label indicator within collapsed final-to-initial glyph-pair
groups. Folios, rather than individual boundaries, are resampled. If a folio is
drawn more than once, all of its observations are duplicated and pair means are
recomputed in the bootstrap sample. The two-way sensitivity estimate
alternately demeans gap and label within glyph-pair and folio groups until
convergence.

\subsection{Blind direct-pixel audit}
\label{app:pixel}

As a stronger check on the bounding-box proxy, we measured the ink-to-ink gap
directly on 700-pixel-wide Beinecke IIIF derivatives of six folios, blind to
the certain/uncertain label. The frozen candidate pool contains 328 eligible
intra-line boundaries (305 certain, 23 uncertain); the sample takes all 23
uncertain and 277 certain boundaries drawn with a fixed NumPy seed
(\texttt{4082026}). Neither the blind manifest nor the blind QC file contains
the label, which resides in a separate key; re-running the measurement
program on the frozen manifest reproduces every stored gap, threshold, and
edge coordinate. Blind QC excluded 14 boundaries (12/277 certain, 2/23
uncertain; Fisher \(p=0.29\)); the two excluded uncertain cases are
algorithmic no-ink failures whose locator-box gaps are 0 and \(-2.55\)
pixels, and each would need an implausible 24-pixel imputed gap to erase the
contrast.

\begin{figure}[tbp]
\centering
\includegraphics[width=\textwidth]{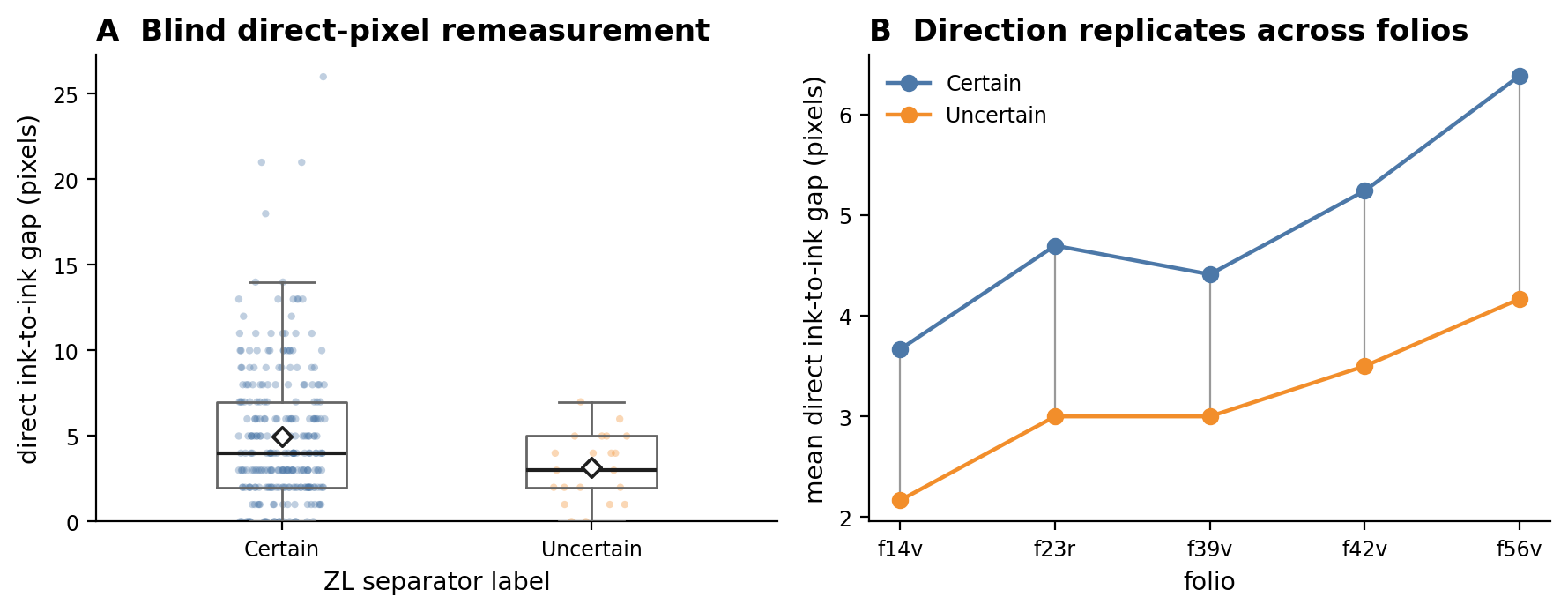}
\caption{Blind direct-pixel validation (700-pixel Beinecke IIIF derivatives).
(A) Direct ink-to-ink gap after blind QC: boxplot with the individual
measurements (median line, mean marked by a white diamond); certain $n=265$
(mean 4.98), uncertain $n=21$ (mean 3.14), pooled certain$-$uncertain
$=+1.84$ px. (B) Mean gap by folio for
the five folios retaining both label types; every folio shows the predicted
sign (5/5; one-sided sign test $p=0.031$).}
\end{figure}

The primary result is 4.98 pixels for certain gaps (median 4, \(n=265\))
against 3.14 pixels for uncertain gaps (median 3, \(n=21\)), a difference of
\(+1.84\) pixels; all five folios retaining both labels have the predicted
sign (exact one-sided sign test \(p=0.031\); equal-folio permutation
preserving each folio's uncertain count \(p=0.021\) one-sided, \(0.066\)
two-sided). The distributions overlap heavily, so the discrimination on the
ink is far weaker than the box AUC of 0.905. The contrast is positive over the
central threshold range (offsets \(-20\) to \(+25\) around the local Otsu
threshold) and under grayscale-Otsu, Lab-lightness, adaptive-Gaussian, and
centre-to-centre-valley estimators, though the last two are weaker; a
vertical-overlap-only stress test produces little separation, so at this
resolution the numerical edge depends on how the vertical support of cursive
strokes is defined. Retaining only measurements whose detected ink edges lie
within 6--12 pixels of the locator-box edges keeps the sign in all five
folios (10-pixel tolerance: 207 certain and 20 uncertain, permutation
\(p=0.052\)). Two post-hoc paired checks agree: the median paired difference
against a nearby certain gap on the same line is \(+2.5\) pixels, and among
the 18 uncertain boundaries with a same-glyph-pair certain comparison the
mean paired contrast is \(+2.05\) pixels, 15/18 in the predicted direction.
The direct gap correlates only modestly with the locator-box gap
(\(r\approx0.23\)), so it is a largely separate measurement, but the same
boxes locate the tokens, so it corroborates that the coordinate contrast is
realised in the ink rather than constituting an independent study.

\begin{figure}[t]
\centering
\includegraphics[width=\textwidth]{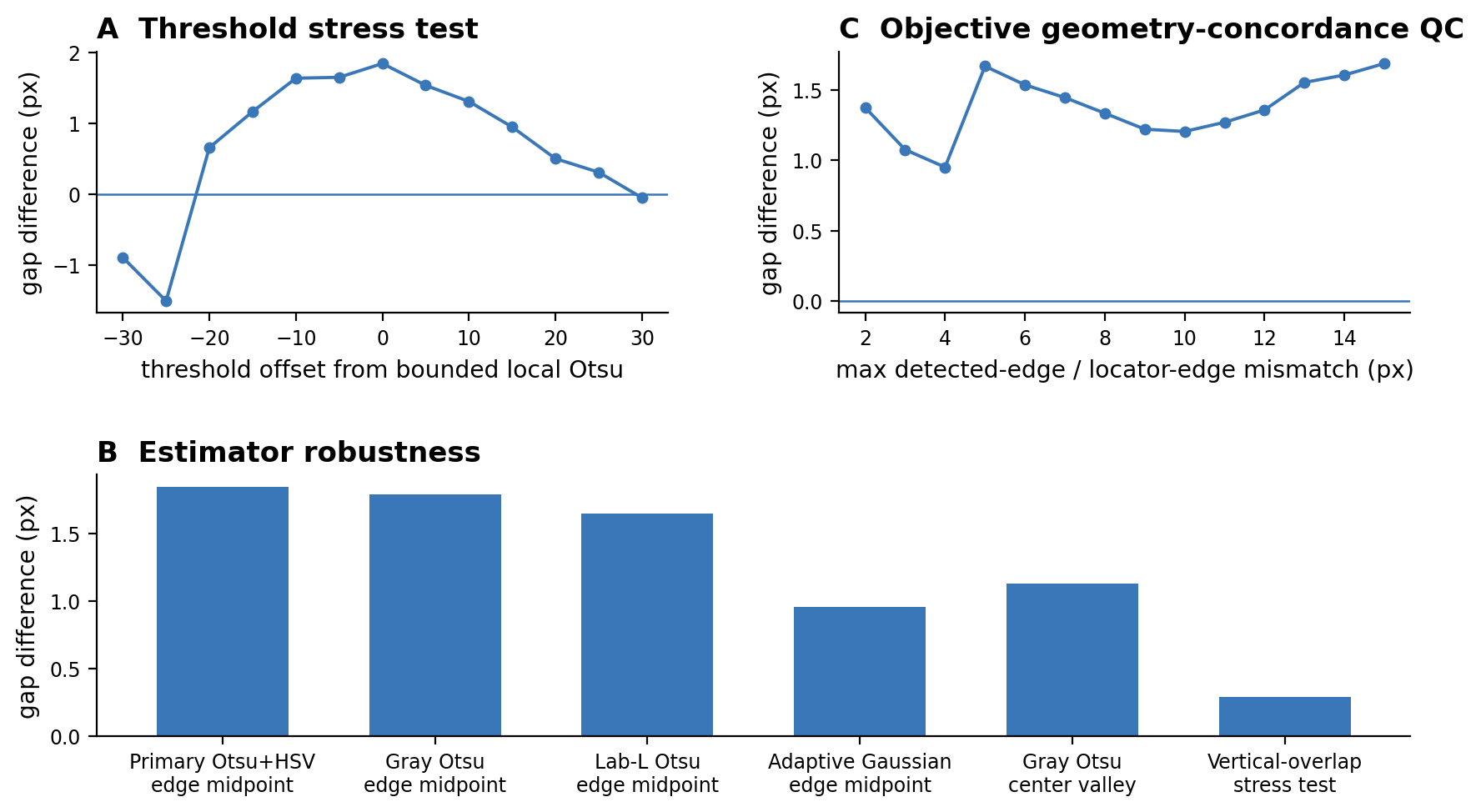}
\caption{Robustness of the blind direct-pixel check. (A) Threshold stress test:
certain$-$uncertain mean gap across grayscale-threshold offsets around the
bounded local Otsu value. (C) Objective geometry-concordance QC: the same
contrast when only boundaries whose detected ink edges lie within a given
tolerance of the locator-box edges are retained. (B) Alternative pixel
estimators; the contrast stays positive for all but the vertical-overlap
stress test.}
\end{figure}

\subsection{Additional robustness detail}
\label{app:robust}

\paragraph{Token order.} At caps of 250, 500, 1,000, 2,000, and 4,000 forms,
the observed-space Voynich shares are 2.35\%, 1.89\%, 1.45\%, 0.79\%, and
\(-0.08\%\); the lowest continuous-text value at the same caps is 3.03\%,
2.77\%, 2.51\%, 2.02\%, and 1.65\%. The high-cap Voynich estimate reaches the
bias floor while the separation remains. Across the 16 leave-one-quire-out
refits the observed-space estimate ranges from 0.48\% to 0.83\% and the
merged-space estimate from 0.23\% to 0.55\%. Currier A and B analysed
separately give \(-0.49\%\) and \(+0.37\%\) at the 2,000 cap (30 shuffles per
stratum); small negative corrected values are finite-sample noise, not
negative mutual information.

\paragraph{Boundary index in raw units.} The within-token mean score is
\(+1.72\) bits, the mean across all intra-line separators \(-2.50\), the
independent-pair anchor \(-2.80\), and the line-break mean \(-2.97\). The
normalisation makes this large negative raw scale readable without changing
the ordering.

\paragraph{Coordinates.} The strict line-by-line alignment retains 17,737
labelled separators with means 0.1272 (certain) and 0.0023 (uncertain). ZL and
v101 provide 11,306 comparable decisions on 1,617 lines; mean normalised gap
is 0.1273 at the 7,905 positions neither marks uncertain, 0.0300 at the 413
marked by one, and \(-0.0044\) at the 62 marked by both (medians 0.1250,
0.0000, 0.0000).

\paragraph{Nesting.} Using the fixed full-corpus association scale, the bootstrap standard error of
mean boundary internality is approximately 9.4 times larger under quire
resampling than under line resampling, corresponding to about 89 times the
variance. The exact factor depends slightly on whether the anchors are held
fixed or recomputed, but the conclusion does not: line-level independence is
not an adequate uncertainty model for corpus-level effects.

The all-quire sign concordance and leave-one-quire-out range provide stronger
small-cluster checks than a line-level \(p\)-value. They show that the effect
is replicated within every observed top-level unit and is not driven by any
one unit. They do not turn 16 manuscript quires into a random sample of
independent manuscripts. The inference is therefore manuscript-wide, not
universal: throughout this codex, uncertain and certain separators behave as
different boundary regimes.

\paragraph{Line length.} Line position within the page is correlated with line length: lines shorten
toward page bottoms, where paragraphs end, and entropy estimated from fewer
glyphs is downward-biased. Without controlling length, an apparent within-page
gradient appears in boundary softness, mean token length, and glyph entropy.
Restricting to the middle 50\% of line lengths (33--54 glyphs; 156 pages,
1,768 lines) removes the gradients: internality \(+0.031\) \([-0.027,+0.082]\),
token length \(-0.003\), entropy \(-0.013\). This is a caution for the wider
line-effects literature; the first-separator and line-initial results reported
here are not generated by this confound, because their contrasts are
constructed within the same line.

\subsection{Calibration of the attack differential and the two further controls}
\label{app:controls}

\paragraph{Differential calibration.} For each synthetic configuration the
cipher was rebuilt with five seeds, an order-3 surrogate was fitted to each
ciphertext, and both were run through the identical pipeline (BPE, Brown
merge to 22 classes, mapping search with eight restarts). Table~\ref{tab:diff}
gives the real-minus-surrogate language-match differential against the
right-language (Latin) and a wrong-language (German) model. Seed 0 of the
three-homophone cipher reproduces the archived single run (\(0.544-0.151=
+0.393\)). Reproduction: \path{decipherment_attack_v6/run_differentials.py}.

\begin{table}[H]
\centering
\small
\caption{Real-minus-surrogate language-match differential for the synthetic
ciphers over five seeds (mean [min, max]).}
\label{tab:diff}
\begin{tabular}{lrr}
\toprule
Configuration & Right language & Wrong language \\
\midrule
H2 (two homophones) & \(+0.21\) [\(+0.09\), \(+0.25\)] & \(+0.09\) [\(+0.05\), \(+0.13\)] \\
H3 (three homophones) & \(+0.33\) [\(+0.11\), \(+0.44\)] & \(+0.09\) [\(+0.05\), \(+0.12\)] \\
H5 (five homophones) & \(+0.30\) [\(+0.23\), \(+0.34\)] & \(+0.10\) [\(+0.09\), \(+0.11\)] \\
H3mix (one-to-three groups) & \(+0.13\) [\(+0.09\), \(+0.22\)] & \(+0.06\) [\(+0.05\), \(+0.10\)] \\
H3rot (rotating tables) & \(+0.52\) [\(+0.45\), \(+0.61\)] & \(+0.10\) [\(+0.06\), \(+0.13\)] \\
Voynichese (14 comparisons) & \multicolumn{2}{c}{\(-0.01\) [\(-0.15\), \(+0.08\)]} \\
\bottomrule
\end{tabular}
\end{table}

\paragraph{Naibbe cipher.} \citet{greshko2025} publishes the cipher tables
and code under a modified MIT licence (repository
\texttt{greshko/naibbe-cipher}, Zenodo 10.5281/zenodo.16415087). Plaintext
spaces are deleted and the letter stream is respaced into single letters
(probability 17/36) and pairs; a single letter becomes one of six
table-specific EVA strings, a pair becomes a prefix string for its first
letter followed by a suffix string for its second (23 letters, six tables,
three states, strings of one to eight glyphs), and the table for each draw is
dealt from a shuffled 52-card deck (\(\alpha\) 20, \(\beta\) 8/8/8, \(\gamma\)
4/4). We transcribed the published table file to
\path{data/controls/naibbe/naibbe_tables.json}, re-implemented the encryption
without executing the repository code, and verified the transcription by
re-deriving all 34,764 tokens of the shipped Pliny sample from its aligned
plaintext (no failures). Our ciphertext encrypts the Caesar text used for the
synthetic calibration (books I--IV; the held-out Latin model is trained on
books V--VIII), prepared as in \path{validate_synthetics.py}, with a fixed
seed: 20,547 words and 123,354 letters become 80,877 tokens and 423,658
glyphs, wrapped to the Voynich line template. Deviations from Greshko's
procedure: the text is respaced as one stream rather than line by line,
\(v\) is merged into \(u\), and his optional 3\% space deletion is not
applied. Statistics use the package functions of the corresponding drivers;
the attack uses the first 120,000 glyphs, 23 classes, eight restarts, and
five seeds, with \path{attack_voynich.ngram_generate} as surrogate. Results
(Caesar; Greshko's Pliny sample in parentheses): \(H\)/\(h_2\) 3.98/2.71 and
3.86/2.34 (3.99/2.71, 3.87/2.34); \(D_k\) 1.555, 1.512, 1.388, 1.158, 1.080,
1.028, 1.160, 1.464, 2.097, 3.016 at \(k=0\ldots1024\) (minimum 1.040 at 64);
82 unit types at \(k=64\), span 2.43; order share 1.25\% (1.06\%) at the
2,000 cap and 1.56/1.48/1.36/1.02\% at 250/500/1,000/4,000; 5,234 types,
40.5\% singletons (5,463, 41.7\%); attack differential against held-out Latin
\(+0.002\) (per seed \(+0.006, -0.001, -0.041, +0.021, +0.038\)) and against
German \(+0.013\) (Pliny: \(-0.008\), \(-0.010\)); at \(k=128\) the
differential moves only to \(+0.01\) to \(+0.06\), equally for the wrong
language; erased-space units cross 2.2\% (1.9\%) of hidden token boundaries
and 7.1\% of the hidden Latin word boundaries. Reproduction:
\path{analysis/reproduce_naibbe_control.py}.

\paragraph{Self-citation generator.} \citet{timm2020} describe a scribe who
copies a nearby glyph group and modifies it by replacing glyphs with similar
ones, adding or deleting glyphs, combining or splitting groups, or duplicating
them, with paragraph-initial gallows and line-initial and line-final
preferences; the reference implementation (Java, MIT, Zenodo
10.5281/zenodo.2531632) fixes the ligature inventory, the similarity and
prefix tables, a curve/line glyph grammar that admits or rejects an edit, the
source chooser (a previous line of the current page, the same writing
position with probability 0.28, paragraph-initial lines drawing from earlier
paragraph-initial lines), and the modification mix (add/remove 0.20,
combine/split 0.30, replace 0.50). We ported these rules to Python
(\path{analysis/reproduce_selfcitation_control.py}) with the Voynich line,
page, and paragraph structure as template; the port at the size of Timm and
Schinner's published sample gives 2,210 types and 62\% singletons against
their 2,228 and 54\%. Two free parameters (an explicit unchanged-copy
probability and the replacement multiplicity) were calibrated to the Voynich
mean token length and vocabulary at 32,747 tokens; the mechanism saturates at
57--60\% singletons for every setting, and its global statistics vary widely
between seeds. Results for three seeds: \(h_2\) 2.64, 2.57, 2.70 collapsed
(2.22, 2.10, 2.22 decomposed); \(D_0\) 1.60--1.63 and \(D_{64}\) 0.649, 0.856,
0.675 (minimum at 64 in every seed, 79--82 units); order share 0.47, 0.59,
0.49\%; 7,125, 4,985, 6,947 types with 60.2, 58.9, 60.4\% singletons;
adjacent tokens within one edit 1.17, 1.06, 1.23 times the shuffle mean;
attack differential \(+0.011, -0.021, +0.018\) (Latin) and \(-0.003, +0.020,
+0.005\) (German); edge-glyph information 0.006--0.009 bits. The crude
glyph-noise surrogate shipped with the bundle reproduces none of these
(\(h_2=3.9\), no trough, zero order), so the agreement is a property of Timm
and Schinner's rules, not of copying as such. With the Java default parameters
(4.5--4.9 thousand types, 56--57\% singletons) the picture is unchanged.

\subsection{Learned units against published inventories}
\label{app:inventory}

Table~\ref{tab:unit_inventory} compares the units learned at 32 and 64 merges
with three published inventories: the 26 multi-glyph ``slot characters'' and
12-slot template of \citet{zattera2022}, the crust--mantle--core word grammar
of \citet{stolfi2000}, and the nine standard EVA composites. Categories are
not exclusive. A string is slot-legal if it segments into slot characters that
can occupy strictly increasing slots of the template; a Stolfi constituent is
a layer letter with its circle modifiers or trailing \(e\) (\texttt{ol},
\texttt{ok}, \texttt{aiin}, \texttt{che}); a fragment is a concatenation of
constituents permitted by his normal-word grammar. Every compound unit
learned by 64 merges is slot-legal (100\% of compound occurrences) and 98\%
of compound occurrences are contiguous paths through Zattera's pruned grammar
(exceptions: \texttt{olk}, \texttt{cthy}, \texttt{ckhy}); but only five
compound units (\texttt{ch}, \texttt{sh}, \texttt{ee}, \texttt{cth},
\texttt{ckh}; 11\% of compound occurrences) are themselves slot characters.
Half of the compound occurrences are two slot characters in non-adjacent
slots (a circle plus the following letter: \texttt{ol}, \texttt{ok},
\texttt{ar}, \texttt{ot}, \texttt{al}, \texttt{or}), the rest span three to
six slots (\texttt{aiin}, \texttt{qok}, \texttt{daiin}, \texttt{chedy},
\texttt{qokeedy}). Against Stolfi's model, two thirds of all occurrences are
exactly one constituent, but a third are fragments that cross his layer
boundaries, chiefly mantle-plus-crust suffixes (\texttt{chedy}, \texttt{edy},
\texttt{eedy}, \texttt{ody}, \texttt{dy}, \texttt{ey}, \texttt{chy},
\texttt{chey}) and the crust-prefix-plus-core group \texttt{qo}, \texttt{qok},
\texttt{qot}. Since both hand-built models were fitted to this corpus and the
slot template admits millions of strings, high coverage is expected; the
informative result is the level. Reproduction:
\path{analysis/reproduce_unit_inventory.py}.

\begin{table}[H]
\centering
\scriptsize
\setlength{\tabcolsep}{4pt}
\caption{Learned BPE units against published word-structure inventories:
shares (\%) of unit types and of unit occurrences in the pooled decomposed-EVA
stream at 32 and 64 merges; the last two columns restrict the 64-merge
inventory to compound units.}
\label{tab:unit_inventory}
\begin{tabular}{lrrrrrr}
\toprule
 & \multicolumn{2}{c}{\(k=32\) (56 types)} & \multicolumn{2}{c}{\(k=64\) (88 types)} & \multicolumn{2}{c}{\(k=64\), compound (63)} \\
\cmidrule(lr){2-3}\cmidrule(lr){4-5}\cmidrule(lr){6-7}
Category & types & occ. & types & occ. & types & occ. \\
\midrule
Single decomposed EVA letter (unmerged) & 44.6 & 38.9 & 28.4 & 27.5 & 0.0 & 0.0 \\
(i) one Zattera slot character & 37.5 & 48.5 & 23.9 & 34.7 & 7.9 & 11.2 \\
(ii) two slot characters, increasing slots & 25.0 & 32.7 & 30.7 & 36.3 & 42.9 & 50.1 \\
\quad of which in adjacent slots & 8.9 & 11.8 & 9.1 & 10.8 & 12.7 & 14.9 \\
(ii\('\)) three or more slot characters & 19.6 & 18.0 & 34.1 & 28.0 & 47.6 & 38.7 \\
Slot-legal string, (i)\(\cup\)(ii)\(\cup\)(ii\('\)) & 82.1 & 99.2 & 88.6 & 99.1 & 98.4 & 100.0 \\
\quad of which a path in Zattera's grammar & 82.1 & 99.2 & 85.2 & 97.3 & 93.7 & 97.6 \\
(iii) one Stolfi layer letter & 33.9 & 33.2 & 21.6 & 25.5 & 7.9 & 11.2 \\
(iii\('\)) one Stolfi constituent & 60.7 & 78.6 & 45.5 & 65.6 & 34.9 & 53.6 \\
(iii\(''\)) legal Stolfi fragment crossing constituents & 23.2 & 19.6 & 44.3 & 33.1 & 61.9 & 45.6 \\
(iv) standard EVA composite & 10.7 & 10.8 & 8.0 & 8.7 & 11.1 & 11.9 \\
(v) none of (i)--(iv) & 14.3 & 0.6 & 9.1 & 0.8 & 1.6 & 0.0 \\
\bottomrule
\end{tabular}
\end{table}

\subsection{Currier strata and line-start detail}
\label{app:dialect}

The ZL page variable gives 1,416 A lines on 114 pages in 11 quires and 2,404 B
lines on 82 pages in 9 quires; 60 lines on unassigned pages are left out, and
a continuation line break belongs to the language of its page. The uncertain
share is the number of ZL commas over all intra-line separators of the
stratum; the raw gap is the mean association score at uncertain minus certain
separators, in bits, on the stratum's own scale. For the line-start
decomposition the line-initial glyph is the first glyph of a line's first
clean token, the mid-line token-initial glyphs are the first glyphs of its
remaining tokens, and enrichment is a glyph's share among line-initial glyphs
divided by its share among mid-line token-initial glyphs of the same line
set. Stratum anchors are within-token \(+1.53\) bits and independent \(-2.30\) bits
for Currier A, and \(+1.91\) and \(-3.08\) for Currier B. Class counts:
A uncertain 781, first 1,393, mid 6,091, certain 8,091, line break 1,283;
B 1,609, 2,381, 15,472, 18,608, 2,307. Independent-stratum quire-bootstrap
90\% intervals for A\,--\,B: uncertain \(-0.009\) \([-0.052,+0.037]\); first
\(+0.036\) \([-0.046,+0.088]\); mid \(+0.088\) \([-0.006,+0.149]\); certain
\(+0.096\) \([-0.001,+0.157]\); line break \(+0.031\) \([-0.054,+0.092]\).
Resampling the 16 quires jointly instead gives \([-0.048,+0.035]\) for the
uncertain difference. Line-start: with paragraph-first lines defined by the
IVTFF paragraph mark (717 versus 3,163 lines), JSD is 0.528, 0.179, and 0.203
(first, other, pooled); the relabelling null is 0.006, 0.002, and 0.001; the
quire-bootstrap 95\% interval of the first-minus-other difference is
\([+0.31,+0.40]\); by language, A 0.51/0.14 and B 0.57/0.24. Enrichments for
\(p,t,k,f,y,d\) are 24.3, 11.9, 3.0, 5.0, 0.38, 0.33 in paragraph-first lines
and 7.9, 3.5, 0.22, 0.65, 5.2, 2.2 elsewhere. Defining the first set instead
as page-opening lines and lines starting a new left-margin unit (ZL locators
\texttt{@} and \texttt{*}, 260 lines) gives JSD 0.487 versus 0.200 and
enrichments 23.2, 17.9, 6.4, 4.6, 0.19, 0.35 versus 17.3, 4.6, 0.46, 2.9, 4.7,
2.0. Reproduction: \path{analysis/reproduce_dialect_linestart.py}.

\subsection{Edge-glyph and token-identity order: full table}
\label{app:edge}

Table~\ref{tab:edgefull} gives the shuffle-corrected mutual information (100
within-line shuffles, seed fixed) for five adjacent-token pairings, each as
excess bits and as a percentage of the marginal entropy of the predicted
feature: (a) last glyph \(\to\) first glyph; (b) last two \(\to\) first two
glyphs; (c) token identity (2,000-type cap) \(\to\) first glyph; (d) last
glyph \(\to\) token identity; (e) token identity \(\to\) token identity (the
statistic of Table~\ref{tab:tokenorder}). ``Destroyed'' is the fraction of the
observed edge-glyph information removed by within-line shuffling; the last
column is the observed-to-shuffled ratio of adjacent tokens within one edit.
Voynich glyphs are composite-collapsed except in the raw-EVA row; the
observed-space Voynich edge information is 0.218 bits before correction, of
which the pair \(y\!\to\!q\) contributes 0.097. Restricting to interior pairs
(dropping line-first and line-last tokens) leaves it at 0.226 bits.
Reproduction: \path{analysis/reproduce_edge_order.py}.

\begin{table}[H]
\centering
\scriptsize
\setlength{\tabcolsep}{4pt}
\caption{Adjacent-token information at five scales (bits/\%).}
\label{tab:edgefull}
\begin{tabular}{lrrrrrrr}
\toprule
Corpus & (a) & (b) & (c) & (d) & (e) & Destroyed & Edit-1 \\
\midrule
Voynich, observed spaces & 0.197/5.7 & 0.243/4.3 & 0.199/5.8 & 0.167/2.0 & 0.067/0.8 & 90\% & 1.12 \\
Voynich, weak spaces merged & 0.162/4.8 & 0.181/3.3 & 0.149/4.4 & 0.121/1.5 & 0.044/0.5 & 88\% & 1.14 \\
Voynich, observed, raw EVA & 0.188/5.9 & 0.233/4.8 & 0.185/5.7 & 0.172/2.1 & 0.067/0.8 & 93\% & 1.13 \\
Voynich, Currier A & 0.137/3.9 & 0.134/2.4 & 0.081/2.3 & 0.045/0.5 & $-0.043$/$-0.5$ & 77\% & 1.20 \\
Voynich, Currier B & 0.248/7.4 & 0.280/5.2 & 0.211/6.3 & 0.191/2.3 & 0.029/0.3 & 91\% & 1.07 \\
Latin narrative & 0.033/0.8 & 0.220/3.4 & 0.174/4.3 & 0.268/3.3 & 0.326/4.0 & 78\% & 0.43 \\
English narrative & 0.076/1.9 & 0.464/7.6 & 0.322/8.2 & 0.339/4.2 & 0.742/9.2 & 81\% & 0.28 \\
French narrative & 0.120/2.9 & 0.469/7.4 & 0.458/11.1 & 0.391/4.9 & 0.782/9.7 & 85\% & 0.53 \\
German narrative & 0.039/0.9 & 0.191/3.1 & 0.188/4.5 & 0.198/2.5 & 0.387/4.8 & 75\% & 0.54 \\
Italian narrative & 0.069/1.7 & 0.336/5.2 & 0.294/7.2 & 0.314/4.0 & 0.483/6.2 & 77\% & 0.54 \\
Latin medical & 0.039/1.0 & 0.191/3.0 & 0.141/3.5 & 0.247/3.0 & 0.295/3.6 & 73\% & 0.29 \\
Latin botanical & 0.018/0.4 & 0.101/1.5 & 0.081/2.0 & 0.126/1.7 & 0.148/2.0 & 65\% & 0.56 \\
English herbal & 0.086/2.1 & 0.470/7.5 & 0.387/9.6 & 0.403/4.8 & 0.794/9.5 & 83\% & 0.47 \\
\emph{Species Plantarum} records & 0.322/7.9 & 1.071/16.4 & 0.850/21.0 & 1.028/12.1 & 1.348/15.8 & 86\% & 2.57 \\
Naibbe cipher (Latin, Caesar) & 0.006/0.2 & 0.038/0.7 & 0.055/1.6 & 0.036/0.4 & 0.107/1.3 & 47\% & 0.82 \\
Self-citation generator, seed 1 & 0.006/0.2 & 0.039/0.7 & 0.017/0.5 & 0.034/0.4 & 0.072/0.9 & 46\% & 1.17 \\
Self-citation generator, seed 2 & 0.009/0.3 & 0.079/1.5 & 0.024/0.8 & 0.058/0.7 & 0.097/1.2 & 60\% & 1.06 \\
Self-citation generator, seed 3 & 0.009/0.3 & 0.050/0.9 & 0.017/0.5 & 0.034/0.4 & 0.054/0.6 & 54\% & 1.23 \\
\bottomrule
\end{tabular}
\end{table}

\subsection{Syllabic-transcription hypothesis: four matched tests}
\label{app:syllable}

The controls are Project Gutenberg texts 26888 (\emph{Bencao Beiyao}) and
23950 (\emph{Three Kingdoms}) converted to pinyin syllable streams with
\texttt{pypinyin} in tonal and toneless variants, split into clauses at
Chinese punctuation (mean 4--5 syllables per clause, shorter than the Voynich
line template, which if anything favours the controls in the shuffle
comparison), and truncated to the Currier-A token count of 10,729. Voynich A
and B here are all paragraph-locus tokens split by the ZL page language
(10,729 and 22,871 tokens; 60 within-line shuffles), a slightly different
subset and shuffle count from the main token-order table, which is why the
Currier estimates differ in the second decimal from those in
Section~\ref{sec:tokenorder}. Table~\ref{tab:syllable} gives the order and
vocabulary comparisons.

\begin{table}[H]
\centering
\small
\caption{Syllabic-transcription hypothesis against genre-matched pinyin
controls at a matched token count. Order share is adjacent-token mutual
information after the within-line shuffle correction, as a percentage of
capped token entropy.}
\label{tab:syllable}
\begin{tabular}{lrrr}
\toprule
Stream & Order share & Types & Hapax \\
\midrule
\emph{Bencao Beiyao} (herbal), tonal      & 5.40\%    & 686     & 20.3\% \\
\emph{Bencao Beiyao} (herbal), toneless   & 5.23\%    & 335     & 9.6\%  \\
\emph{Three Kingdoms} (narrative), tonal  & 6.21\%    & 749     & 17.6\% \\
\emph{Three Kingdoms} (narrative), toneless & 6.42\%  & --      & --     \\
Voynichese, Currier A                     & $-0.53\%$ & 3{,}343 & 72.2\% \\
Voynichese, Currier B (22,871 tokens)     & $+0.41\%$ & 4{,}840 & 68.6\% \\
\bottomrule
\end{tabular}
\end{table}

Two supplementary checks are less specific. First, under within-token BPE the
toneless pinyin \emph{letter} stream (131,030 letters) has a dependence gap of
1.40 bits at zero merges, close to the Voynich 1.60, but rises to 1.58 bits by
64 merges, whereas the Voynich gap falls to 1.05; the letters of a phonetic
syllabary do not show the early trough. Since plain Latin also rises while
verbose Latin ciphers trough, this is a scale signature, not a language test.
Second, running the calibrated attack of Section~\ref{sec:units} with a
trigram model trained on the concatenated pinyin letters gives language-match
indices of 0.105 (Currier A) and 0.126 (Currier B) against surrogate values of
0.138 and 0.117: a differential of approximately zero, compared with the
\(+0.39\) that a genuine cipher shows over its own surrogate. This is a test
of Voynichese as a homophonic substitution of pinyin letters, not of the
token-as-syllable reading itself. The manuscript's vellum is radiocarbon-dated
to 1404--1438; the statistical result does not rely on any chronological
argument. Reproduction: \path{decipherment_attack_v10/syllabic_test.py}.

\subsection{Reproduction}

All drivers take their input roots as explicit arguments, fix random seeds,
and accept \texttt{--json-output}; the README lists the exact commands and
smoke-test flags. Defaults are 1,500 quire/folio bootstrap draws, 300
matched-length windows per entropy control, 100 shuffles, 100 random quire
partitions, 50 randomisations, and five cipher seeds. NumPy and Matplotlib are
cited as research software \citep{harris2020,hunter2007}.

\footnotesize
\setlength{\bibsep}{1pt}


\begin{thebibliography}{99}

\bibitem[Amancio et al.(2013)]{amancio2013}
Amancio, D. R., Altmann, E. G., Rybski, D., Oliveira, O. N., Jr., and Costa,
L. da F. 2013.
\newblock Probing the statistical properties of unknown texts: application to
the Voynich manuscript.
\newblock \emph{PLOS ONE} 8(7):e67310.
\newblock \url{https://doi.org/10.1371/journal.pone.0067310}.

\bibitem[Beinecke Library(n.d.)]{beinecke}
Beinecke Rare Book and Manuscript Library. n.d.
\newblock The Beinecke Cipher (Voynich) Manuscript, MS 408.
\newblock \url{https://beinecke.library.yale.edu/beinecke/collections/beinecke-cipher-voynich-manuscript}.

\bibitem[Bennett(1976)]{bennett1976}
Bennett, W. R., Jr. 1976.
\newblock \emph{Scientific and Engineering Problem-Solving with the Computer}.
\newblock Prentice-Hall, Englewood Cliffs, NJ.

\bibitem[Bowern and Gaskell(2022)]{bowern2022}
Bowern, C. and Gaskell, D. E. 2022.
\newblock Enciphered after all? Word-level text metrics are compatible with
some types of encipherment.
\newblock In \emph{Proceedings of the International Conference on the Voynich
Manuscript 2022}, CEUR Workshop Proceedings 3313, paper 6.
\newblock \url{https://ceur-ws.org/Vol-3313/paper6.pdf}.

\bibitem[Bowern and Lindemann(2021)]{bowern2021}
Bowern, C. and Lindemann, L. 2021.
\newblock The linguistics of the Voynich manuscript.
\newblock \emph{Annual Review of Linguistics} 7:285--308.
\newblock \url{https://doi.org/10.1146/annurev-linguistics-011619-030613}.

\bibitem[Brown et al.(1992)]{brown1992}
Brown, P. F., Della Pietra, V. J., deSouza, P. V., Lai, J. C., and Mercer, R. L.
1992.
\newblock Class-based \emph{n}-gram models of natural language.
\newblock \emph{Computational Linguistics} 18(4):467--480.
\newblock \url{https://aclanthology.org/J92-4003/}.

\bibitem[Caruana et al.(2022)]{caruana2022}
Caruana, A., Layfield, C., and Abela, J. 2022.
\newblock An analysis of the relationship between words within the Voynich
manuscript.
\newblock In \emph{Proceedings of the International Conference on the Voynich
Manuscript 2022}, CEUR Workshop Proceedings 3313, paper 8.
\newblock \url{https://ceur-ws.org/Vol-3313/paper8.pdf}.

\bibitem[Church and Hanks(1990)]{church1990}
Church, K. W. and Hanks, P. 1990.
\newblock Word association norms, mutual information, and lexicography.
\newblock \emph{Computational Linguistics} 16(1):22--29.
\newblock \url{https://aclanthology.org/J90-1003/}.

\bibitem[Claston(n.d.)]{v101}
Claston, G. n.d.
\newblock Voynich manuscript transcription, version 101.
\newblock \url{https://www.voynich.nu/data/voyn_101.txt}.

\bibitem[Coffee et al.(2013)]{coffee2013}
Coffee, N., Koenig, J.-P., Poornima, S., Forstall, C. W., Ossewaarde, R., and
Jacobson, S. L. 2013.
\newblock The Tesserae Project: Intertextual analysis of Latin poetry.
\newblock \emph{Literary and Linguistic Computing} 28(2):221--228.
\newblock \url{https://doi.org/10.1093/llc/fqs033}.

\bibitem[Cohen(1960)]{cohen1960}
Cohen, J. 1960.
\newblock A coefficient of agreement for nominal scales.
\newblock \emph{Educational and Psychological Measurement} 20(1):37--46.
\newblock \url{https://doi.org/10.1177/001316446002000104}.

\bibitem[Cover and Thomas(2006)]{cover2006}
Cover, T. M. and Thomas, J. A. 2006.
\newblock \emph{Elements of Information Theory}, 2nd ed.
\newblock Wiley-Interscience, Hoboken, NJ.
\newblock \url{https://doi.org/10.1002/047174882X}.

\bibitem[Currier(1976)]{currier1976}
Currier, P. H. 1976.
\newblock Some important new statistical findings; further details; and
Voynich manuscript: Some notes and observations.
\newblock In M. E. D'Imperio (ed.), \emph{New Research on the Voynich
Manuscript: Proceedings of a Seminar, 30 November 1976}, 20--27, 45--65.
\newblock \url{https://media.defense.gov/2021/Jul/13/2002761428/-1/-1/0/PROCEEDINGS-OF-A-SEMINAR-30-NOVEMBER-1976.PDF}.

\bibitem[D'Imperio(1978)]{dimperio1978}
D'Imperio, M. E. 1978.
\newblock \emph{The Voynich Manuscript: An Elegant Enigma}.
\newblock National Security Agency/Central Security Service, Fort George G.
Meade, MD.

\bibitem[Efron and Tibshirani(1993)]{efron1993}
Efron, B. and Tibshirani, R. J. 1993.
\newblock \emph{An Introduction to the Bootstrap}.
\newblock Chapman \& Hall, New York.

\bibitem[Fagin Davis(2020)]{fagindavis2020}
Fagin Davis, L. 2020.
\newblock How many glyphs and how many scribes? Digital paleography and the
Voynich manuscript.
\newblock \emph{Manuscript Studies} 5(1):164--180.
\newblock \url{https://doi.org/10.1353/mns.2020.0011}.

\bibitem[Fawcett(2006)]{fawcett2006}
Fawcett, T. 2006.
\newblock An introduction to ROC analysis.
\newblock \emph{Pattern Recognition Letters} 27(8):861--874.
\newblock \url{https://doi.org/10.1016/j.patrec.2005.10.010}.

\bibitem[Feaster(2020)]{feaster2020}
Feaster, P. 2020.
\newblock Ruminations on the Voynich manuscript.
\newblock \emph{Griffonage-Dot-Com}, 24 August 2020.
\newblock \url{https://griffonagedotcom.wordpress.com/2020/08/24/ruminations-on-the-voynich-manuscript/}.

\bibitem[Feaster(2021)]{feaster2021}
Feaster, P. 2021.
\newblock Transitional probabilities in the Voynich manuscript.
\newblock \emph{Griffonage-Dot-Com}, 20 September 2021.
\newblock \url{https://griffonagedotcom.wordpress.com/2021/09/20/transitional-probabilities-in-the-voynich-manuscript/}.

\bibitem[Gaskell and Bowern(2022)]{gaskell2022}
Gaskell, D. E. and Bowern, C. 2022.
\newblock Gibberish after all? Voynichese is statistically similar to
human-produced samples of meaningless text.
\newblock In \emph{Proceedings of the International Conference on the Voynich
Manuscript 2022}, CEUR Workshop Proceedings 3313, paper 4.
\newblock \url{https://ceur-ws.org/Vol-3313/paper4.pdf}.

\bibitem[Greshko(2025)]{greshko2025}
Greshko, M. A. 2025.
\newblock The Naibbe cipher: a substitution cipher that encrypts Latin and
Italian as Voynich Manuscript-like ciphertext.
\newblock \emph{Cryptologia}, published online 26 November 2025.
\newblock \url{https://doi.org/10.1080/01611194.2025.2566408}.

\bibitem[Harris et al.(2020)]{harris2020}
Harris, C. R. et al. 2020.
\newblock Array programming with NumPy.
\newblock \emph{Nature} 585:357--362.
\newblock \url{https://doi.org/10.1038/s41586-020-2649-2}.

\bibitem[Hauer and Kondrak(2016)]{hauer2016}
Hauer, B. and Kondrak, G. 2016.
\newblock Decoding anagrammed texts written in an unknown language and script.
\newblock \emph{Transactions of the Association for Computational Linguistics}
4:75--86.
\newblock \url{https://doi.org/10.1162/tacl_a_00084}.

\bibitem[Hunter(2007)]{hunter2007}
Hunter, J. D. 2007.
\newblock Matplotlib: A 2D graphics environment.
\newblock \emph{Computing in Science \& Engineering} 9(3):90--95.
\newblock \url{https://doi.org/10.1109/MCSE.2007.55}.

\bibitem[Landini(2001)]{landini2001}
Landini, G. 2001.
\newblock Evidence of linguistic structure in the Voynich manuscript using
spectral analysis.
\newblock \emph{Cryptologia} 25(4):275--295.
\newblock \url{https://doi.org/10.1080/0161-110191889932}.

\bibitem[Lindemann and Bowern(2021)]{lindemann2021}
Lindemann, L. and Bowern, C. 2021.
\newblock Character entropy in modern and historical texts: comparison metrics
for an undeciphered manuscript.
\newblock arXiv:2010.14697.
\newblock \url{https://arxiv.org/abs/2010.14697}.

\bibitem[Matlach et al.(2022)]{matlach2022}
Matlach, V., Jane\v{c}kov\'{a}, B. A., and Dost\'{a}l, D. 2022.
\newblock The Voynich manuscript: symbol roles revisited.
\newblock \emph{PLOS ONE} 17(1):e0260948.
\newblock \url{https://doi.org/10.1371/journal.pone.0260948}.

\bibitem[Montemurro and Zanette(2013)]{montemurro2013}
Montemurro, M. A. and Zanette, D. H. 2013.
\newblock Keywords and co-occurrence patterns in the Voynich manuscript: an
information-theoretic analysis.
\newblock \emph{PLOS ONE} 8(6):e66344.
\newblock \url{https://doi.org/10.1371/journal.pone.0066344}.

\bibitem[Paninski(2003)]{paninski2003}
Paninski, L. 2003.
\newblock Estimation of entropy and mutual information.
\newblock \emph{Neural Computation} 15(6):1191--1253.
\newblock \url{https://doi.org/10.1162/089976603321780272}.

\bibitem[Parisel(2025)]{parisel2025}
Parisel, C. 2025.
\newblock Directionality of the Voynich script.
\newblock arXiv:2509.10573.
\newblock \url{https://arxiv.org/abs/2509.10573}.

\bibitem[Parisel(2026a)]{parisel2026a}
Parisel, C. 2026a.
\newblock Evidence of layered positional and directional constraints in the
Voynich manuscript: implications for cipher-like structure.
\newblock arXiv:2604.19762.
\newblock \url{https://arxiv.org/abs/2604.19762}.

\bibitem[Parisel(2026b)]{parisel2026b}
Parisel, C. 2026b.
\newblock A quantitative confirmation of the Currier language distinction.
\newblock arXiv:2604.25979.
\newblock \url{https://arxiv.org/abs/2604.25979}.

\bibitem[Perseus Digital Library(n.d.)]{perseus}
Perseus Digital Library. n.d.
\newblock Greek and Roman materials, edited by G. R. Crane.
\newblock Tufts University.
\newblock \url{https://www.perseus.tufts.edu/}.

\bibitem[Project Gutenberg(n.d.)]{gutenbergcontrols}
Project Gutenberg. n.d.
\newblock Electronic texts used as controls: Caesar, ebooks 218, 18837, and
10657; Verne, 5097; Grimmelshausen, 55171; Manzoni, 45334; Murri, 26166;
\emph{Bollettino del Club Alpino Italiano}, 27608; Culpeper, 49513; and
Linnaeus, 20771 and 27049.
\newblock \url{https://www.gutenberg.org/}.

\bibitem[Ravi and Knight(2011)]{ravi2011}
Ravi, S. and Knight, K. 2011.
\newblock Bayesian inference for Zodiac and other homophonic ciphers.
\newblock In \emph{Proceedings of the 49th Annual Meeting of the Association
for Computational Linguistics: Human Language Technologies}, 239--247.
\newblock \url{https://aclanthology.org/P11-1025/}.

\bibitem[Reddy and Knight(2011)]{reddy2011}
Reddy, S. and Knight, K. 2011.
\newblock What we know about the Voynich manuscript.
\newblock In \emph{Proceedings of the 5th ACL-HLT Workshop on Language
Technology for Cultural Heritage, Social Sciences, and Humanities}, 78--86.
\newblock \url{https://aclanthology.org/W11-1511/}.

\bibitem[RobGea(2022)]{robgea2022}
RobGea. 2022.
\newblock Bigrams across uncertain spaces.
\newblock \emph{The Voynich Ninja} forum, thread 3803, 21 May 2022.
\newblock \url{https://www.voynich.ninja/thread-3803.html}.

\bibitem[Rugg(2004)]{rugg2004}
Rugg, G. 2004.
\newblock An elegant hoax? A possible solution to the Voynich manuscript.
\newblock \emph{Cryptologia} 28(1):31--46.
\newblock \url{https://doi.org/10.1080/0161-110491892755}.

\bibitem[Schinner(2007)]{schinner2007}
Schinner, A. 2007.
\newblock The Voynich manuscript: evidence of the hoax hypothesis.
\newblock \emph{Cryptologia} 31(2):95--107.
\newblock \url{https://doi.org/10.1080/01611190601133539}.

\bibitem[Sefaria(n.d.)]{sefaria}
Sefaria. n.d.
\newblock Sefaria API: Texts endpoint and structured library of Jewish texts.
\newblock \url{https://developers.sefaria.org/reference/get-v3-texts}.

\bibitem[Sennrich et al.(2016)]{sennrich2016}
Sennrich, R., Haddow, B., and Birch, A. 2016.
\newblock Neural machine translation of rare words with subword units.
\newblock In \emph{Proceedings of the 54th Annual Meeting of the Association
for Computational Linguistics}, 1715--1725.
\newblock \url{https://aclanthology.org/P16-1162/}.

\bibitem[Smith and Ponzi(2019)]{smith2019}
Smith, E. M. and Ponzi, M. 2019.
\newblock Glyph combinations across word breaks in the Voynich manuscript.
\newblock \emph{Cryptologia} 43(6):466--485.
\newblock \url{https://doi.org/10.1080/01611194.2019.1596998}.

\bibitem[Steckley and Steckley(2024)]{steckley2024}
Steckley, A. and Steckley, N. 2024.
\newblock Subtle signs of scribal intent in the Voynich manuscript.
\newblock arXiv:2404.13069.
\newblock \url{https://arxiv.org/abs/2404.13069}.

\bibitem[Stolfi(1997)]{stolfi1997chinese}
Stolfi, J. 1997.
\newblock The generalized Chinese theory.
\newblock Web note, 23 November 1997.
\newblock \url{https://www.ic.unicamp.br/~stolfi/EXPORT/00-EXPORT/97-11-23-tonal/}.

\bibitem[Stolfi(2000)]{stolfi2000}
Stolfi, J. 2000.
\newblock A grammar for Voynichese words.
\newblock Web note, 14 June 2000.
\newblock \url{https://www.ic.unicamp.br/~stolfi/EXPORT/00-EXPORT/00-06-07-word-grammar/}.

\bibitem[Stolfi(2002)]{stolfi2002}
Stolfi, J. 2002.
\newblock Chinese theory redux: comparing the VMS and East Asian word length
distributions.
\newblock Web note, 18 January 2002.
\newblock \url{https://www.ic.unicamp.br/~stolfi/EXPORT/00-EXPORT/02-01-18-chinese-redux/}.

\bibitem[Timm and Schinner(2020)]{timm2020}
Timm, T. and Schinner, A. 2020.
\newblock A possible generating algorithm of the Voynich manuscript.
\newblock \emph{Cryptologia} 44(1):1--19.
\newblock \url{https://doi.org/10.1080/01611194.2019.1596999}.

\bibitem[Vogt(2012)]{vogt2012}
Vogt, E. 2012.
\newblock The line as a functional unit in the Voynich manuscript: some
statistical observations.
\newblock Working paper, 27 November 2012.
\newblock \url{https://voynichthoughts.files.wordpress.com/2012/11/the_voynich_line.pdf}.

\bibitem[Voynich Ninja(2020)]{voynichninja2020}
Voynich Ninja. 2020.
\newblock Vord length distribution.
\newblock \emph{The Voynich Ninja} forum, thread 3261, 28 June 2020.
\newblock \url{https://www.voynich.ninja/thread-3261.html}.

\bibitem[Voynichese(n.d.)]{voynichese}
Voynichese. n.d.
\newblock Query viewer and per-token folio-layout coordinates for the Voynich
manuscript.
\newblock \url{https://voynichese.com/}.

\bibitem[Zandbergen and Landini(n.d.)]{zl3b}
Zandbergen, R. and Landini, G. n.d.
\newblock ZL transliteration of the Voynich manuscript, version 3b, IVTFF 2.0.
\newblock \url{https://voynich.nu/}.

\bibitem[Zattera(2022)]{zattera2022}
Zattera, M. 2022.
\newblock A new transliteration alphabet brings new evidence of word structure
and multiple ``languages'' in the Voynich manuscript.
\newblock In \emph{Proceedings of the International Conference on the Voynich
Manuscript 2022}, CEUR Workshop Proceedings 3313, paper 10.
\newblock \url{https://ceur-ws.org/Vol-3313/paper10.pdf}.

\bibitem[Zipf(1949)]{zipf1949}
Zipf, G. K. 1949.
\newblock \emph{Human Behavior and the Principle of Least Effort: An
Introduction to Human Ecology}.
\newblock Addison-Wesley, Cambridge, MA.

\end{thebibliography}
\end{document}